\documentclass{article}

\PassOptionsToPackage{round}{natbib}
\usepackage[preprint]{neurips_2026}

\usepackage[utf8]{inputenc}
\usepackage[T1]{fontenc}
\usepackage{hyperref}
\usepackage{url}
\usepackage{booktabs}
\usepackage{multirow}
\usepackage{makecell}
\usepackage{amsfonts}
\usepackage{amsmath}
\usepackage{amssymb}
\usepackage{mathtools}
\usepackage{amsthm}
\usepackage{nicefrac}
\usepackage{microtype}
\usepackage{xcolor}
\usepackage{graphicx}
\usepackage{subcaption}
\usepackage{wrapfig}
\usepackage[capitalize,noabbrev]{cleveref}
\usepackage{tikz}
\usetikzlibrary{positioning, arrows.meta, calc, shapes.geometric}
\usepackage{pifont}
\usepackage{enumitem}

\colorlet{black70}{black!70}

\DeclareMathOperator*{\argmin}{arg\,min}

\newcommand{\csq}[1]{%
  \begingroup
  \setlength{\fboxsep}{0pt}%
  \setlength{\fboxrule}{0.4pt}%
  \fcolorbox{#1}{#1!30}{\rule{0pt}{0.5em}\rule{0.5em}{0pt}}%
  \endgroup
}

\theoremstyle{plain}

\theoremstyle{definition}

\theoremstyle{remark}

\title{Neural Field Tokenizations with Hierarchy and Spatial Locality Priors}

\author{%
  Alonso Urbano\textsuperscript{1}\thanks{Corresponding author: \texttt{urbano@zib.de}.}\quad
  David W.\ Romero\textsuperscript{2}\quad
  Max Zimmer\textsuperscript{1}\quad
  Sebastian Pokutta\textsuperscript{1,3} \\[3pt]
  \textsuperscript{1}\,Department for AI in Society, Science, and Technology, Zuse Institute Berlin (ZIB), Germany \\
  \textsuperscript{2}\,Cartesia AI, San Francisco, CA, USA \\
  \textsuperscript{3}\,Institute of Mathematics, Technische Universität Berlin, Germany \\[2pt]
  \texttt{\{urbano,zimmer,pokutta\}@zib.de}\quad\texttt{dwromerog@gmail.com}
}
\begin{document}

\maketitle

%%% PAPER BODY

\begin{abstract}
Neural fields parameterize data as functions from coordinates to values, providing a unified framework for representation learning across modalities. Existing approaches are dominated by per-sample meta-learning, which scales poorly due to memory-intensive inner-loop optimization. The natural alternative --feed-forward encoding-- typically introduces modality-specific assumptions, sacrificing the generality that makes learning with neural fields attractive. We argue that \emph{locality} and \emph{hierarchy} are useful priors for learning field representations that can be injected without compromising modality-agnosticism. We propose LH-NeF, a framework to learn general-purpose tokenized representations of continuous signals. A locality-preserving hierarchical encoder maps raw coordinate-value field observations to structured tokens, from which the field is reconstructed during training. By replacing meta-learning's inner loop with a single forward pass, LH-NeF uses 42× less memory and supports 133× larger batches than the strongest modality-agnostic baseline. Across images, 3D shapes, and climate fields, our learned representations match or exceed performance of modality-agnostic, modality-specific, and specialized generative neural field baselines on both reconstruction and downstream tasks.
\end{abstract}

\vspace{-3mm}
\section{Introduction}
\vspace{-2mm}

\begin{figure}[t]
\centering
\begin{tikzpicture}[
    >=Stealth,
    every node/.style={inner sep=0pt},
    arr/.style={->, semithick, color=black70},
    lbl/.style={font=\scriptsize, text=black!80},
    scale=0.82, every node/.append style={transform shape},
]

% === ROW 1: Tokenizer ===

% --- Row label ---
\node[font=\sffamily\footnotesize, text=black!80] at (-6.8, 0) {(a)};

% --- Side modalities (tiny thumbnails, left, stacked) ---
\node[anchor=east] (era5) at (-4.75, 0.15) {%
    \includegraphics[height=0.75cm]{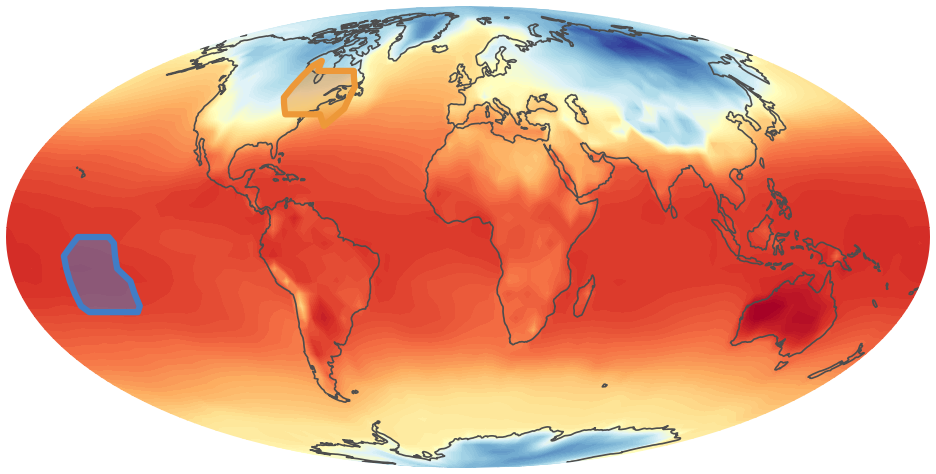}};
\node[anchor=east] (shapenet) at (-4.15, -0.45) {%
    \includegraphics[height=1.35cm]{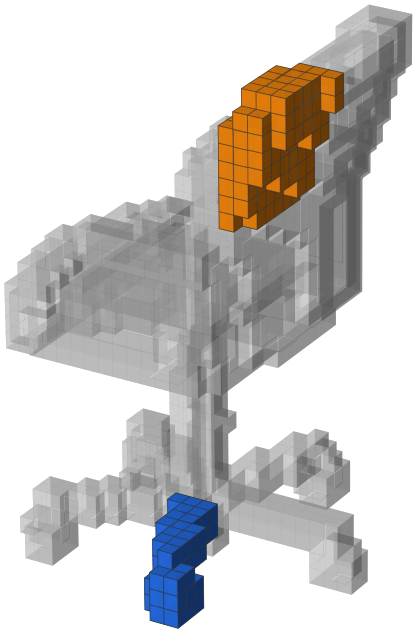}};
% "Any modality" label — adjust (x, y) to position freely
\node[font=\scriptsize, text=black!80] at (-5.2, 1.05) {\shortstack{Any modality \\$f : \mathcal{X} \to \mathbb{R}^{C_{\mathrm{out}}}$}};

% --- Dashed separator ---
\draw[densely dashed, black!30, thin] (-3.85, 0.8) -- (-3.85, -0.8);

% --- Input: CelebA main image ---
\node (celeba) at (-2.4, 0) {%
    \includegraphics[height=2.4cm]{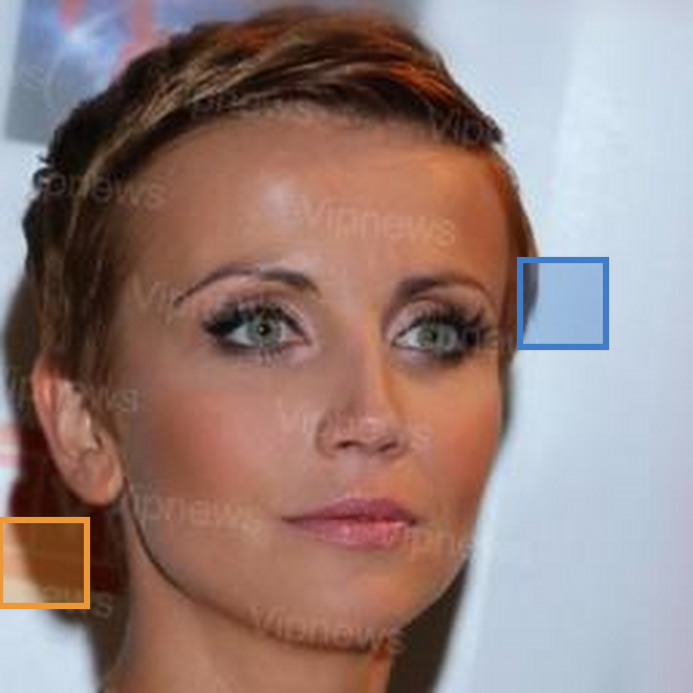}};
% Observations label — below the image
%\node[font=\scriptsize, text=black!75] at (-2.4, -1.42) {$\{(x_i, v_i)\}_{i=1}^N$};

% --- Arrow: input -> tokens ---
\draw[arr] (-1.1, 0) -- (2.8, 0);
% Arrow label — short, above arrow
\node[lbl, align=center] at (0.8, 0.35) {\shortstack{Embed observations $\{(x_i, v_i)\}$\\+ Locality-preserving re-order}};

% --- Tokens ---
\node (tokens) at (3.5, 0) {%
    \includegraphics[height=2.0cm]{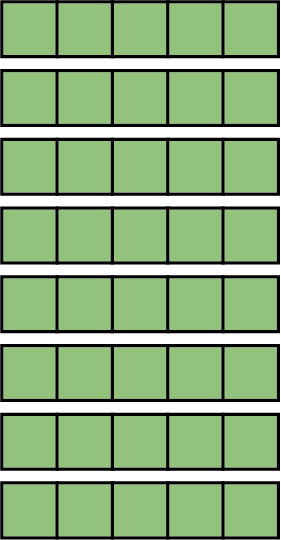}};
\node[lbl] at (3.5, -1.3) {$N \!\times\! C_{\mathrm{emb}}$};

% --- Arrow: tokens -> HiP ---
\draw[arr] (4.2, 0) -- (4.7, 0);

% --- HiP Encoder block (tikz inline) ---
% shift = position (left/right, up/down), scale = size
\begin{scope}[shift={(5.9, 0)}, scale=0.75, every node/.append style={transform shape}]
\input{figures/main_figure/hip_block_overlay.tex}
\end{scope}
\node[lbl] at (5.9, -0.85) {$L$ blocks};

% --- Arrow: HiP -> grouped rep ---
\draw[arr] (7.1, 0) -- (7.7, 0);

% --- Grouped token representation ---
\node (groups) at (8.3, 0) {%
    \includegraphics[height=2.0cm]{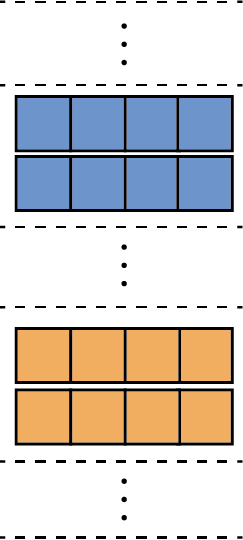}};
% Label above: full name
\node[lbl, font=\scriptsize] at (8.22, 1.3) {Grouped tokens $\mathbf{Y}^{(L)}$};
% Label below: math — Y with dimensions
\node[lbl] at (8.2, -1.3) {$G_L \times K_L \times C_L$};

% === ROW 2: Renderer ===

% --- Row label ---
\node[font=\sffamily\footnotesize, text=black!80] at (-6.8, -3.2) {(b)};

% --- 4 groups panel (leftmost) ---
\node (fourgroups) at (-4.6, -3.2) {%
    \scalebox{.9}{\input{figures/main_figure/routing_overlay.tex}}};

% --- Arrow: routing -> cross-attn ---
\draw[arr] (-3., -3.2) -- (-2.3, -3.2);

% --- Cross-attention blocks (4 groups) ---
\node at (2.3, -3.35) {%
    \resizebox{8.9cm}{!}{%
        \begin{tikzpicture}[>=Stealth]
        \input{figures/main_figure/xattn_overlay.tex}
        \end{tikzpicture}%
    }};

% --- FiLM + output stack ---
\node at (9.0, -3.1) {%
    \resizebox{!}{2.8cm}{%
        \begin{tikzpicture}[>=Stealth]
        \input{figures/main_figure/film_overlay.tex}
        \end{tikzpicture}%
    }};

% --- Arrow: h(x) → FiLM (connecting the two sub-figures) ---
% Three segments: right, up, right. Adjust these 4 values:
%   start_x=2.3  start_y=-4.5  turn_x=5.5  end_x=6.2  end_y=-3.1
\draw[arr, thin] (2.9, -4.34) -- (7.8, -4.34) -- (7.8, -2.75) -- (8.34, -2.75);

% --- Weight equation (compact: softmax of Gaussian) ---
\node[lbl, font=\tiny] at (-4.5, -4.7) {%
    $w_g(x) = \mathrm{softmax}\!\bigl({-}d_{\mathcal{X}}(x, \mu_g)^2 / 2\sigma_\theta^2\bigr)$};

\end{tikzpicture}
\vspace{-1mm}
\caption{LH-NeF overview.
(a) Observations from an input are embedded into the $C_{\mathrm{emb}}$-dimensional tokenizer input space (Appx.~\ref{app:encoder}) and sorted by a locality-preserving ordering, then processed by $L$ grouped attention blocks, yielding the tokenized representation $\mathbf{Y}^{(L)}$. The locality-preserving ordering ensures each group's spatial support covers a compact region of the coordinate space (\csq{blue!60!cyan}\,/\,\csq{orange}).
(b) To render the field $f_\theta$ at any coordinate $x$, the renderer routes $x$ to the $k$ nearest groups, weighting each group's contribution via a learnable Gaussian kernel. Per-group cross-attention outputs are then aggregated into $\mathbf{h}(x)$ via a weighted sum, FiLM-conditioned on the query's relative coordinate $\tilde{x}$, and decoded into the field value $f_\theta(x)$.
\vspace{-3mm}
}
\label{fig:main_figure}
\end{figure}

Neural fields represent data as continuous functions from coordinates to values parameterized by neural networks that can be queried at arbitrary resolution, e.g.\ an image as a mapping $f_\theta: [0,1]^2 \to \mathbb{R}^3$. Learning over entire datasets is enabled by conditional neural fields, where a shared network $f_\theta(x;z)$ is conditioned on latent variables $z$ representing each instance in the data~\citep{park2019deepsdf,mescheder2019occupancy}. While many methods focused on specific modalities (e.g. 2D grids or radiance fields), \citet{dupont2022data} proposed \emph{Functa}, a unified framework to learn latent representations $z\in\mathbb{R}^n$ of neural fields \emph{across modalities}. The resulting latents are then used as data surrogates for arbitrary downstream tasks such as generation and classification~\citep{bauer2023spatial,wessels2024equivariant}.
%This modality-agnostic interface for learning representations has since found applications across fields such as medical imaging~\citep{friedrich2026medfuncta}, video~\citep{wolleb2025vidfuncta}, and PDEs~\citep{jo2025pdefuncta, wessels2024enf_pde}.
However, existing modality-agnostic neural field methods face a fundamental tradeoff between introducing useful structure\break into these representations and preserving the modality-agnosticism that makes them attractive.

%In standard representation learning, structure is often introduced through encoder design: for instance, convolutional networks leverage translation equivariance and local receptive fields~\citep{lecun1998gradient}, Vision Transformers (ViT) partition inputs into compact local regions (patches) processed by self-attention~\citep{dosovitskiy2020vit}, and graph neural networks propagate information along edges via message passing~\citep{kipf2017gcn}. 
In standard representation learning, structure is introduced through encoder design, e.g., local receptive fields in CNN~\citep{lecun1998gradient}, patch tokenization in ViT~\citep{dosovitskiy2020vit}, message passing in GCN~\citep{kipf2017gcn}.
These design choices shape the latent space in ways that improve efficiency and generalization~\citep{bronstein2021geometric}, but typically assume specific input structure, which constrains the modality-agnosticism of neural fields when used to obtain the conditioning latent $z$. As a result, current modality-agnostic neural field methods (Functa and its descendants) avoid obtaining latents through structured encoders entirely, and resort to decoder-only setups where the latents are obtained per-sample through meta-learning~\mbox{\citep{finn2017maml,dupont2022data}} or auto-decoding~\citep{park2019deepsdf}. While effective, this reliance on per-sample optimization has practical and theoretical costs. 

First, the resulting optimization-derived latent space lacks explicit structure, which makes learning\break inefficient. Recent methods address this by introducing structure through geometry-grounded mechanisms in the decoding process~\citep{wessels2024equivariant} or modality-specific latent layouts~\citep{bauer2023spatial}. However, these approaches remain tied to optimization-based latent fitting and in many instances sacrifice modality-agnosticism at the cost of adding structure --see Table~\ref{tab:method_comparison} for an overview.
Second, meta-learning approaches --the dominant paradigm on state-of-the-art methods-- require storing the full computation graph in memory across all inner steps to compute second-order gradients. At moderate resolutions, this severely limits batch sizes even at moderate resolutions (Table~\ref{tab:efficiency}), a major bottleneck for scaling to high resolutions or large datasets. Overall, current approaches face a trade-off between (i) structural priors that aid learning, (ii) modality-agnosticism, and (iii) the scalability bottleneck of training with per-sample optimization.

We argue that \emph{hierarchy} (fine-to-coarse organization) and \emph{locality} (correlation between nearby coordinates) are inductive biases that can be used to \emph{learn continuous signal tokenizations} without compromising modality-agnosticism --both have in fact been proved to be excellent priors across images~\citep{lecun1998gradient,vahdat2020nvae}, 3D geometry~\citep{qi2017pointnetpp,wang2023octformer} and continuous signals in general~\citep{mallat1989multiresolution}. To this end, we propose LH-NeF (\underline{L}ocality-preserving \underline{H}ierarchical \underline{Ne}ural \underline{F}ields). LH-NeF consists of a \emph{tokenizer} that builds upon general perception systems exploiting hierarchy through layered grouped attention (Hierarchical Perceiver~\citep{carreira2022hip}), and modifies them to respect coordinate-space locality. As a result, the spatial support of each group corresponds to a compact region in the domain (Figure~\ref{fig:groups_compact}). To render field values from the resulting tokenization, LH-NeF defines a \emph{renderer} that exploits this structure through Gaussian soft group routing and group-wise cross-attention.
The same architecture handles any coordinate domain, requiring only a locality-preserving ordering for the domain geometry, and infers neural field representations in a single forward pass --eliminating the memory bottleneck that limits the scalability of existing modality-agnostic neural field methods.

Our contributions are:
\begin{itemize}[leftmargin=*,itemsep=1pt,topsep=2pt,parsep=0pt]
    \item We identify a tradeoff between scalability, structural priors and modality-agnosticism in neural field representation learning, and show that hierarchy and locality can be introduced without modality-specific architectural assumptions.
    \item We propose LH-NeF, a modality-agnostic framework for learning neural field representations. The LH-NeF \emph{tokenizer} produces a spatially structured grouped tokenization that conditions a \emph{renderer}, which decodes the field at any coordinate via soft group routing and cross-attention aggregation.
    \item We demonstrate empirically that LH-NeF stays competitive or exceeds state-of-the-art reconstruction and downstream performance on images, 3D shapes, and climate data, while using ${\sim}42{\times}$ less memory and fitting $133{\times}$ larger batch sizes than the strongest modality-agnostic baseline.
\end{itemize}

Our code is publicly available at \texttt{link-hidden-for-double-blind-review}.

\begin{table}[t]
\centering
\caption{Methods for learning on datasets of neural fields. OPT ${=}$ optimization-based. $^\dagger$\,labels used during meta-learning.}
\label{tab:method_comparison}
\setlength{\aboverulesep}{1.5pt}
\setlength{\belowrulesep}{1.5pt}
\resizebox{\textwidth}{!}{%
\footnotesize
\begin{tabular}{@{}lllll@{}}
\toprule
Method & Latent structure & Latent inference & Modality & Objective \\
\midrule
GEM \citep{du2021gem} & Vector & OPT (autodec.) & Any & \multirow{3}{*}{Generation} \\
GASP \citep{dupont2021gasp} & Vector & --- & Any &  \\
DPF \citep{zhuang2023dpf} & --- & --- & Any &  \\
MWT \citep{gielisse2025endtoend} & Weights & OPT (MAML) & Any & Classif.$^\dagger$ \\
NF2Vec \citep{nf2vec2023} & Vector & OPT (fit+enc.) & 3D & \multirow{5}{*}{Repr.\ learn} \\
3DS2VS \citep{zhang2023shape2vecset} & Vector set & Forward pass & 3D &  \\
Spatial Functa \citep{bauer2023spatial} & 2D vector grid & OPT (MAML) & 2D &  \\
Functa \citep{dupont2022data} & Vector & OPT (MAML) & Any &  \\
ENF \citep{wessels2024equivariant} & Point cloud & OPT (MAML) & Any &  \\
\midrule
LH-NeF (Ours) & Structured tokens & Forward pass & Any & Repr.\ learn \\
\bottomrule
\end{tabular}}
\vspace{-3mm}
\end{table}

\section{Related work}
\label{sec:related}
\vspace{-2mm}

\paragraph{Neural fields and conditional neural fields.}
Neural fields parameterize signals as continuous functions $f_\theta: \mathbb{R}^d \to \mathbb{R}^c$ via coordinate-based neural networks.
Early works fit a separate network per signal (NeRF~\citep{mildenhall2020nerf}, SIREN~\citep{sitzmann2020siren}), with acceleration via hash grids~\citep{mueller2022instant} and other spatial structures~\citep{chen2022tensorf,barron2021mipnerf}.
To extend to datasets, DeepSDF~\citep{park2019deepsdf} introduced Conditional Neural Fields with auto-decoding: a single shared MLP conditioned on per-shape latent codes optimized jointly with network weights.
Functa~\citep{dupont2022data} extended this paradigm to learning modality-agnostic representations for downstream tasks.
This paradigm has been widely adopted, e.g., for medical imaging~\citep{friedrich2026medfuncta}, video~\citep{wolleb2025vidfuncta,wolleb2025lrmfuncta}, PDEs~\citep{jo2025pdefuncta, wessels2024enf_pde}, and generation~\citep{du2021gem,dupont2021gasp,zhuang2023dpf} --with diffusion models trained on conditioning variable spaces~\citep{peebles2023dit,ho2020ddpm,chen2024infd,kim2023neuralfield_ldm}.
All modality-agnostic methods in this family rely on per-sample optimization (MAML or auto-decoding) to obtain latents.
Several works observe that the resulting latent vectors are difficult to classify directly~\citep{wolleb2025vidfuncta,gielisse2025endtoend}, and \citet{gielisse2025endtoend} showed that more fitting steps improve reconstruction but hurt classification.
Recent modality-agnostic works improve performance by adding structure: 2D latent grids~\citep{bauer2023spatial}, geometry-grounded point clouds with symmetry-specific bi-invariants and Gaussian-weighted conditioning~\citep{wessels2024equivariant} or weight-space embeddings directly~\citep{nf2vec2023,navon2023dws,gielisse2025endtoend}.
Akin to our method, 3DShape2VecSet~\citep{zhang2023shape2vecset} encodes shapes as unstructured latent vector sets, but lacks spatial structure and is modality-specific. At query time, several local conditioning methods read from multiple features via interpolation or nearest neighbors on feature grids~\citep{chen2021liif, lee2022lte, yu2021pixelnerf, peng2020convocc}. Our renderer applies the same principle to modality-agnostic grouped tokens under a locality-preserving ordering.

\vspace{-2mm}
\paragraph{Forward-pass inference of neural field representations.}
Encoder-based methods that bypass per-sample optimization of the field representation exist, but typically in modality-specific settings.
An early example is Occupancy Networks~\citep{mescheder2019occupancy}, which pair a PointNet/ResNet encoder with a shared occupancy decoder; amortized, but restricted to 3D shapes.
In 2D, LIIF~\citep{chen2021liif} and LTE~\citep{lee2022lte} pair CNN encoders with local coordinate-conditioned MLP decoding, conditioning each query on its nearest grid features with distance-based weighting. INFD~\citep{chen2024infd} uses a CNN encoder for neural field diffusion and leverages neural field properties to obtain state-of-the-art multi-scale generation.
In 3D, pixelNeRF~\citep{yu2021pixelnerf} and LRM~\citep{hong2024lrm} condition radiance/tri-plane representations on image features with camera models. ConvOccNet~\citep{peng2020convocc} uses trilinear interpolation from volumetric feature grids and 3DILG~\citep{zhang20223dilg} conditions on irregular latent point sets with Gaussian-weighted aggregation. For PDEs, AROMA~\citep{serrano2024aroma} uses a Perceiver-style encoder. In general, these methods are either tied to a specific input modality or lack a more general structure that enables learning across modalities.
LH-NeF bridges this gap. It encodes the field in grouped tokens by operating on the field's coordinate-value observations, without relying on modality-specific components.
Outside of neural fields literature, modality-agnostic models exist. Perceiver IO~\citep{jaegle2021perceiverio} uses cross-attention to learn in a modality agnostic way. However, its learned latent bottleneck lacks spatial structure and locality. Building on Perceiver IO, Hierarchical Perceiver (HiP)~\citep{carreira2022hip} adds hierarchy but violates locality, which our ablations find to significantly hurt performance for data defined on metric spaces that carry a corresponding geometry (Table~\ref{tab:ablation}). These methods learn representations in modality-agnostic settings, but can not query inputs at arbitrary resolutions as neural fields allow. 

\vspace{-2mm}
\paragraph{Hierarchy, locality, and dynamic tokenization.}
Hierarchy and locality are fundamental inductive biases in deep learning.
CNNs learn fine-to-coarse features through stacked layers of local filters with increasing receptive fields~\citep{lecun1998gradient}, ViTs introduce locality through explicit patch-based tokenization before applying global self-attention~\citep{dosovitskiy2020vit}.
Swin Transformer~\citep{liu2021swin} combines both via hierarchical shifted windows over image patches and
HiP~\citep{carreira2022hip} generalizes hierarchical processing to arbitrary modalities via grouped hierarchical attention. However, HiP's grouping follows a fixed raster scan that ignores coordinate-space geometry, violating locality. Our approach respects locality and produces tokenized representations whose geometry is input-adaptive and locality-preserving (\cref{fig:voronoi_3d}). The benefits of hierarchy extend beyond continuous signals.
 Recently, H-Net~\citep{hwang2025hnet} used dynamic chunking to replace fixed tokenization with input-adaptive hierarchical grouping and showed that learned hierarchical representations improve efficiency in large language models.
Other work on dynamic (or adaptive) tokenization addresses this in vision settings.
TokenLearner~\citep{ryoo2021tokenlearner} selects tokens via spatial attention, ALIT~\citep{duggal2025alit} and ElasticTok~\citep{yan2025elastictok} adapt token count to input complexity, and GPSToken~\citep{zhang2025gpstoken} parameterizes tokens as 2D Gaussians. However, all these methods assume fixed modalities (text and images, respectively).
LH-NeF combines input-adaptive spatial grouping with hierarchical attention and exploits this structure explicitly at decoding time, creating spatially coherent receptive fields at every level without grid assumptions or modality-specific windowing (Appx.~\ref{app:hierarchy}).

\vspace{-3mm}
\section{Method}
\label{sec:method}
\vspace{-2mm}

We consider the problem of representing a signal as a continuous field $f_\theta : \mathcal{X} \to \mathbb{R}^{C_{\mathrm{out}}}$, where the coordinate domain $(\mathcal{X}, d_{\mathcal{X}})$ is a metric space of dimension $d=\mathrm{dim(\mathcal{X})}$ equipped with a metric $d_{\mathcal{X}}$ (e.g., for images, $\mathcal{X} = [-1,1]^2$ with the Euclidean distance; for climate data, $\mathcal{X} = S^2$ with the Riemannian distance).
Given $N$ observations $\{(x_i, v_i)\}_{i=1}^N$ of $f$ with coordinates $x_i \in \mathcal{X}$ and values $v_i = f(x_i) \in \mathbb{R}^{C_{\mathrm{out}}}$, our goal is to learn a (conditional) neural field $f_\theta$ that enables querying the input at any coordinate $x \in \mathcal{X}$.

\vspace{-2mm}
\paragraph{Overview.}
Our method has two main components (Fig.~\ref{fig:main_figure}).
First, the LH-NeF \emph{tokenizer} takes observations $\{(x_i,v_i)\}_{i=1}^N$ of a field $f$ and produces a \emph{grouped token representation} $\mathbf{Y}^{(L)}$ with spatial metadata (\S\ref{sec:encoder}), representing the field $f$.
Then, we define a conditional neural field: given a tokenization $\mathbf{Y}^{(L)}$ of $f$ and a query coordinate $x$, we route $x$ to the spatially relevant groups of $\mathbf{Y}^{(L)}$ and aggregate information via cross-attention to decode the field value $f_\theta(x)$ (\S\ref{sec:renderer}). The pipeline is trained end-to-end by minimizing $\ell_1$ reconstruction loss between $f_\theta(x)$ and the ground truth $f(x)$.

\vspace{-2mm}
\subsection{LH-NeF Tokenizer}
\label{sec:encoder}
\begin{wrapfigure}[15]{r}{0.32\textwidth}
\vspace{-12pt}
\centering

\resizebox{0.9\linewidth}{!}{%
\begin{tikzpicture}[>=Stealth]
\input{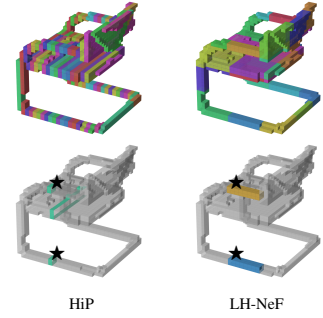}
\end{tikzpicture}%
}\\[1pt]
{\tiny \hspace{1.2em} HiP \hspace{8em} LH-NeF}
\vspace{-3pt}
\caption{Group assignments on a 3D chair.
Top: all groups.
Bottom: routed groups for two queries ($\bigstar$).}
\label{fig:groups_compact}
\end{wrapfigure}

Our tokenizer builds on Hierarchical Perceivers (HiP)~\citep{carreira2022hip}, which process input sequences by splitting tokens into contiguous groups, applying attention within each group, and merging groups into coarser ones across blocks.
The original formulation operates on generic token sequences with a fixed (flattened) rasterization order.
In the neural field setting, each element of the sequence of field observations carries a coordinate $x_i\in\mathcal{X}$ that lives on a metric space $(\mathcal{X}, d_{\mathcal{X}})$. As a result, the contiguous grouping of HiP leads to each group's coordinates occupying a spatial region of $\mathcal{X}$ over which attention exchanges information -- essentially \emph{receptive fields}.
Under HiP's default raster ordering, these regions induce \emph{slices} over $\mathcal{X}$, e.g., consecutive rows of pixels in a 2D image, or slices in a 3D voxelization as shown in \cref{fig:groups_compact} (left). This violates the spatial locality of $(\mathcal{X}, d_{\mathcal{X}})$, as coordinates that are far in $\mathcal{X}$ share the same token group. Notably, the initial sequence ordering determines each group's spatial support and, by extension, the geometry of the entire downstream hierarchy (Figure~\ref{fig:hierarchy_cifar10}).

Building on this observation, we reorder input sequences before grouping using a locality-preserving permutation $\pi:\{1,...,N\}\to\{1,...,N\}$ derived from \emph{space-filling curves}~\citep{bader2013space} on the input coordinates.
This reordering provides contiguous groups that correspond to spatially compact regions on on $\mathcal{X}$ rather than slices (\csq{blue!60!cyan}\,/\,\csq{orange} in \cref{fig:main_figure}a), and allows locality to propagate through the hierarchy by merging nearby groups into progressively coarser spatial regions -- see visualizations in Appx.~\ref{app:hierarchy}.

\textbf{Locality-preserving permutation $\pi$.}
We define a locality-preserving key $\kappa:\mathcal{X}\to\mathbb{N}$ and sort tokens by ascending $\kappa(x_i)$.
For Euclidean domains $\mathcal{X}\subseteq\mathbb{R}^d$, we quantize coordinates $x_i \in [-1,1]^d$ into a discrete grid of $2^b$ bins per dimension, and use Morton ordering (bit-interleaving) as the locality key $\kappa$ (Appx.~\ref{app:orderings_euclidean}). Other Euclidean locality-preserving keys apply equally (Table~\ref{tab:ablation}). For non-Euclidean domains, we can construct locality-preserving keys based on transformed Hilbert curves~\citep{AI2025101899}, which are defined for any Riemannian manifold -- e.g., we use $S^2$ cell indices for the sphere (Appx.~\ref{app:ordering_domains}). The permutation $\pi$ is then defined by sorting tokens in ascending order of $\kappa(x_i)$. This yields an ordering where nearby coordinates remain close in the reordered sequence (\cref{fig:groups_compact}).

After reordering, we embed the sequence as
$e_i = \mathrm{Proj}_v(v_{\pi(i)}) + \mathrm{PE}(x_{\pi(i)})$,
where $\mathrm{Proj}_v$ is a learned projection and $\mathrm{PE}$ is a sinusoidal positional encoding (\citet{vaswani2017attention}) on $d$ dimensions.
We then form groups by taking contiguous chunks of this reordered sequence and process it through $L$ grouped attention blocks just like in HiP.
We denote by $\mathcal{I}^{(\ell)}_g \subseteq \{1,...,N\}$ the set of (reordered) sequence indices that fall into group $g$ at block $\ell$.

\textbf{Grouped tokens with spatial metadata.}
The output of each block $\ell$ is a set of grouped tokens $\mathbf{Y}^{(\ell)} {=} \{\mathbf{Y}^{(\ell)}_g\}_{g=1}^{G_\ell}$, $\mathbf{Y}^{(\ell)}_g \in \mathbb{R}^{K_\ell \times C_\ell}$. Note that each group $g$ is a spatially local subset of the \emph{observed} coordinate-value input pairs and is therefore input-adaptive. The final block output $\mathbf{Y}^{(L)}$ is the tokenized representation of the field. In contrast to standard HiP, which collapses onto a single group at the final block ($G_L{=}1$), we retain multiple groups at the last block to control the granularity of the resulting tokenization (e.g., $G_L{=}32$). We summarize the spatial support of groups $g\in\{1,...,G_L\}$ at the final block by their centroid $\mu_g$ and extent $\lambda_g$:
{\setlength{\abovedisplayskip}{4pt}\setlength{\belowdisplayskip}{1pt}
\begin{equation}
\mu_g = \argmin_{y\in\mathcal{X}}
  \sum_{i \in \mathcal{I}_g} d_{\mathcal{X}}(y, x_i)^2, \qquad
\lambda_g = \max_{i \in \mathcal{I}_g} d_{\mathcal{X}}(\mu_g, x_i),
\end{equation}}
where $\mathcal{I}_g=\mathcal{I}^{(L)}_g$ denotes the indices assigned to group $g$ at the final encoder block. The centroid is the Fréchet mean of the group's observed coordinates (the arithmetic mean on Euclidean domains $\mathcal{X}\subseteq\mathbb{R}^d$), and the extent is the group radius. On Euclidean domains, we replace the scalar radius with the axis-aligned bounding-box extent
$\lambda_g = \max_i x_i - \min_i x_i \in \mathbb{R}^d_{>0}$, which provides richer information that is not available on general metric spaces. Together with $\mathbf{Y}^{(L)}$, the pairs $\{(\mu_g, \lambda_g)\}_{g=1}^{G_L}$ (\emph{routing metadata}) condition the field renderer as we explain next.
% centroids $\mu_g$ determine which groups are nearest to a query (and their routing weights), while extents $\lambda_g$ provide a local length scale that normalizes the query's relative offset within a group.

\vspace{-2mm}
\subsection{LH-NeF Renderer}
\label{sec:renderer}
Given conditioning variables $(\mathbf{Y}^{(L)}, \{(\mu_g, \lambda_g)\}_{g=1}^{G_L})$ of $f$ and a query coordinate $x \in \mathcal{X}$, the LH-NeF renderer produces the field value $f_\theta(x)$ in three steps: \emph{(i)} \textit{routing} identifies which token groups are spatially relevant to $x$, and assign soft weights to them, \emph{(ii)} \textit{aggregation} uses cross-attention to read out information of each relevant group and \emph{(iii)} \textit{modulation} combines per-group outputs and applies a geometry-conditioned modulation to produce the final field value $f_\theta(x)$.% \& output (combine the per-group outputs, apply a geometry-conditioned modulation, and produce the field value $f_\theta(x)$).

\textbf{(i) $k$-Group soft routing.}
We identify the $k$ groups whose centroids are nearest to $x$:
\begin{equation}
\mathcal{N}(x) = \argmin_{S \subseteq \{1,\ldots,G\}, |S|=k} \sum_{g \in S} d_{\mathcal{X}}(x, \mu_g)^2,
\end{equation}
according to the distance $d_{\mathcal{X}}$ in $\mathcal{X}$, and each selected group gets a weight assigned via a Gaussian kernel with learnable bandwidth $\sigma_\theta \in \mathbb{R}_{>0}$:
\begin{equation}
w_g(x) = \frac{\exp\left(-d_{\mathcal{X}}(x, \mu_g)^2 / 2\sigma_\theta^2\right)}{\sum_{g' \in \mathcal{N}(x)} \exp\left(-d_{\mathcal{X}}(x, \mu_{g'})^2 / 2\sigma_\theta^2\right)}, \quad g \in \mathcal{N}(x).
\end{equation}
The motivation for multi-group routing is \emph{continuity}. With hard routing ($k{=}1$), queries would switch abruptly from one group to another as a query moves across the boundary of their receptive fields. This would cause potential discontinuities in the reconstructed field.
Multi-group routing ensures $f_\theta$ changes \emph{smoothly} across $\mathcal{X}$. 

\begin{table}[t]
\centering
\caption{Reconstruction quality across modalities.
Test set PSNR (dB, $\uparrow$) for images, IoU ($\uparrow$) for 3D shapes, MSE ($\downarrow$) for climate temperatures, and field parameter count (\#Param, $\downarrow$). \textbf{Bold}: best.}
\label{tab:reconstruction}
\footnotesize
\setlength{\tabcolsep}{3.75pt}
\begin{tabular*}{\textwidth}{@{\extracolsep{\fill}}l cc cc cc cc cc@{}}
\toprule
& \multicolumn{6}{c}{Images} & \multicolumn{2}{c}{3D Shapes} & \multicolumn{2}{c}{Data on Manifolds} \\
\cmidrule(lr){2-7}\cmidrule(lr){8-9}\cmidrule(lr){10-11}
& \multicolumn{2}{c}{CIFAR-10} & \multicolumn{2}{c}{CelebA-HQ $64^2$} & \multicolumn{2}{c}{ImageNet1k $256^2$} & \multicolumn{2}{c}{ShapeNet16} & \multicolumn{2}{c}{ERA5} \\
\cmidrule(lr){2-3}\cmidrule(lr){4-5}\cmidrule(lr){6-7}\cmidrule(lr){8-9}\cmidrule(lr){10-11}
Method & PSNR$\uparrow$ & \#Param & PSNR$\uparrow$ & \#Param & PSNR$\uparrow$ & \#Param & IoU$\uparrow$ & \#Param & $T_t$-MSE$\downarrow$ & \#Param \\
\midrule
Functa          & 38.1 & 2.6M & 28.0 & 3.4M$^\ddagger$ & -- & -- & 92.1 & 4.0M$^\ddagger$ & 5.75E-05 & 4.1M$^\ddagger$ \\
Spatial Functa$^\dagger$ & 39.0 & 425K$^\ddagger$ & -- & -- & \textbf{38.4}\,/\,28.3$^\S$ & 1.4M$^\ddagger$ & -- & -- & -- & -- \\
ENF             & 42.2 & 522K & 34.6 & 3.2M$^\ddagger$ & 27.5 & \textbf{817K}$^\ddagger$ & 92.9 & 813K$^\ddagger$ & \textbf{8.04E-06} & 817K$^\ddagger$ \\
\midrule
LH-NeF (Ours)  & \textbf{44.4}\scalebox{0.7}{$\pm0.3$} & \textbf{198K} & \textbf{36.1}\scalebox{0.7}{$\pm0.2$} & \textbf{657K} & 28.3 & 1.8M & \textbf{93.1}\scalebox{0.7}{$\pm0.2$} & \textbf{649K} & 3.57E-05 & \textbf{307K} \\
\bottomrule
\end{tabular*}
\\[3pt]
{\scriptsize
$^\dagger$\,2D-grid data only, non modality-agnostic.\quad
$^\ddagger$\,Not reported; reproduced from provided code (cf.\ Appx.~\ref{app:baselines}).\\[1pt]
$^\S$\,Spatial Functa uses 65K conditioning dimensions vs.\ ${\leq}$14K for ENF\,/\,LH-NeF; at comparable latent budget (16K), they report 28.3\,dB.
}
\vspace{-3mm}
\end{table}

While bounded receptive fields already provide coarse locality (each group only ``sees'' tokens from a spatial region), Gaussian weighting adds \emph{distance-dependent relevance} during inference: groups whose centroids are farther away from the queried coordinate contribute less.
The bandwidth $\sigma_\theta$ defines a characteristic scale --the contribution of groups beyond $\sim 2 \sigma_\theta$ away from the query becomes negligible. We learn $\sigma_\theta$ during training and parameterize it as $\sigma_\theta = \exp(\log\sigma_\theta)$ to ensure positivity.
% Overall, this provides an extra, fine-grained locality bias at rendering time. 

\textbf{(ii) Cross-attention aggregation.}
We embed the query coordinate into a query vector:
\begin{equation}
\mathbf{q}(x) = \mathrm{MLP}_q\big(\mathrm{PE}(x)\big) \in \mathbb{R}^D
\end{equation}
and use cross-attention to extract relevant information from the $k$-nearest groups according to the queried coordinate $x$. In particular, for each routed group $g \in \mathcal{N}(x)$, we apply cross-attention with $\mathbf{Y}_g$ as keys/values:
\begin{equation}
\mathbf{h}_g(x) = \mathrm{CrossAttn}\big(\mathbf{q}(x),\; \mathbf{Y}_g\big) \in \mathbb{R}^D.
\end{equation}
Each group's contribution is then aggregated into a feature vector via a weighted sum:
\begin{equation}
\mathbf{h}(x) = \sum_{g \in \mathcal{N}(x)} w_g(x) \cdot \mathbf{h}_g(x).
\end{equation}

\textbf{(iii) Geometry-conditioned modulation.}
Next, we use FiLM-style modulations~\citep{perez2018film} to condition $\mathbf{h}(x)$ based on the position of the query $x$ \emph{within} its nearest group. In general metric spaces, the query's extent-normalized position relative to its nearest group center $\mu_{g^*}$ is:
\begin{equation}
\tilde{x} = \frac{\log_{\mu_{g^*}}(x)}{\lambda_{g^*}},
\end{equation}
where $g^* {=} \argmin_{g \in \mathcal{N}(x)} d_{\mathcal{X}}(x, \mu_g)$ is the nearest group, and $\log_{\mu_{g^*}}{:} \mathcal{X} {\to} T_{\mu_{g^*}}\mathcal{X}$ is the logarithmic map mapping $x$ into the tangent space at $\mu_{g^*}$ (which reduces to $x - \mu_{g^*}$ on Euclidean spaces). This normalized coordinate $\tilde{x}$ then modulates $\mathbf{h}(x)$ as:
\begin{equation}
(\gamma, \beta) = \mathrm{MLP}_{\mathrm{FiLM}}\big(\mathrm{PE}(\tilde{x})\big), \qquad
\mathbf{h}'(x) = (1 + \gamma) \odot \mathbf{h}(x) + \beta,
\end{equation}
which lets the renderer adapt its output based on intra-group position (e.g.\ near vs.\ far from group boundaries). The final field value is
\begin{equation}
f_\theta(x) = \mathrm{MLP}_{\mathrm{out}}(\mathbf{h}'(x)) \in \mathbb{R}^{C_{\mathrm{out}}}.
\end{equation}

\begin{table}[t]
\centering
\caption{Downstream tasks on frozen latent representations.
\textbf{Bold}: best among modality-agnostic representation learning methods. \underline{Underline}: best overall.}
\label{tab:downstream}
\small
\scalebox{0.88}{
\begin{tabular*}{\textwidth}{@{\extracolsep{\fill}}l cc cc c@{}}
\toprule
& \multicolumn{2}{c}{Generation (FID$\downarrow$)} & \multicolumn{2}{c}{Classification (Acc$\uparrow$)} & Forecasting (MSE$\downarrow$) \\
\cmidrule(lr){2-3}\cmidrule(lr){4-5}\cmidrule(lr){6-6}
Method & CIFAR-10 & CelebA-HQ $64^2$ & CIFAR-10 & ShapeNet16 & ERA5 \\
\midrule
GEM$^\ddagger$  & 23.8 & 30.4  & -- & -- & -- \\
GASP$^\ddagger$ & --   & 13.5  & -- & -- & -- \\
DPF$^\ddagger$  & \underline{15.1} & 13.2  & -- & -- & -- \\
\midrule
Spatial Functa$^\dagger$ & 16.5 & --    & 90.3\% & -- & -- \\
\midrule
Functa          & 78.2 & 40.4  & 68.3\% & 90.3\% & 3.45E-03 \\
ENF             & 23.5 & 33.8  & 82.1\% & 96.6\% & \textbf{\underline{9.44E-06}} \\
\midrule
LH-NeF (Ours)  & \textbf{18.5} & \textbf{\underline{9.7}} & \textbf{\underline{91.2\%}} &  \textbf{\underline{96.8\%}} & 4.42E-05 \\
\bottomrule
\end{tabular*}}
\\[3pt]
{\scriptsize $^\dagger$\,2D-grid data only, non modality-agnostic.\quad $^\ddagger$\,Generative only, non general repr.\ learning.}
\vspace{-0.8\baselineskip}
\end{table}

\vspace{-2mm}
\subsection{LH-NeF properties}
\vspace{-1mm}

The design choices above give rise to several properties worth highlighting:%, quantified in \S\ref{sec:ablations} where applicable:
\begin{itemize}[leftmargin=*,itemsep=2pt,topsep=1pt,parsep=0pt]
\item \emph{Locality} enters at three scales: the tokenizer's coordinate-ordered grouping summarizes each input into spatially compact regions; Gaussian-weighted routing at query time blends information from the $k$ nearest groups; and FiLM modulation provides finer adaptation based on where within the nearest group's receptive field the query lies. Because $\tilde{x}$ is group-normalized, this modulation is invariant to translations and axis-aligned rescalings of the group's support, automatically sharing the same conditioning across groups (Appx.~\ref{app:film_invariance}, visualized in \cref{fig:film_tiling}).

\item \emph{Hierarchy.}
The tokenizer's layered blocks induce a multi-scale hierarchy that is aligned with coordinate-space geometry; e.g., on dense grids, successive layers produce progressively coarser compact spatial tilings (Appendix~\ref{app:hierarchy} visualizes this for 2D inputs and 3D shapes).

\item \emph{Dynamic tokenization.}
Because grouping operates \emph{over observations} rather than over the coordinate space itself, the group support $(\mu_g,\lambda_g)$ adapts to the sampling density and geometry of the input. For signals with uniform sampling (e.g., images, voxelized data), the support of groups is consistent for all instances. For signals with non-uniform sampling however (e.g. point clouds), our tokenization adapts to the geometry of each instance (see \cref{fig:voronoi_3d}).

\item \emph{Generality.}
Our framework applies to any modality expressed as coordinate-value observations over a coordinate domain $\mathcal{X}$. That is, the natural scope of neural fields. The method only requires distinguishing between Euclidean and non-Euclidean domains to provide a correct locality key.

\item \emph{Efficiency.}
Our approach replaces second-order MAML's $\mathcal{O}(K \cdot \mathrm{Mem}(\nabla\mathcal{L}))$ inner-loop cost~\citep{rajeswaran2019imaml} with a single forward-backward pass, yielding ${\sim}42{\times}$ memory reduction and $133{\times}$ larger batch sizes in practice at $64{\times}64$ resolution (\S\ref{sec:ablations}). At inference, $k$-group routing maintains querying costs at $\mathcal{O}(k K_L D)$ rather than $\mathcal{O}(G K_L D)$ (often $k \in \{2,4\}$).
\end{itemize}

\vspace{-3mm}
\section{Experiments}
\label{sec:experiments}
\vspace{-2mm}

We evaluate LH-NeF across three modalities: 2D images (CIFAR-10 $32^2$, CelebA-HQ $64^2$, ImageNet1k $256^2$), 3D voxel occupancy (ShapeNet16, $32^3$), and global climate fields (ERA5). The coordinate dimension $d$ and locality-preserving key $\kappa$ are set by the corresponding coordinate domain. Other hyperparameters are chosen based on best validation loss (full configurations in Appx.~\ref{app:stage1}).

We follow the usual two-stage pipeline: \textit{(i)} we fit LH-NeF to the data through a reconstruction objective, and then \textit{(ii)} use the learned tokenization, obtained through a forward pass on the frozen LH-NeF tokenizer, to solve downstream tasks. We compare against three functa-style neural field representation learning methods:
Functa~\citep{dupont2022data},
Spatial Functa~\citep{bauer2023spatial}
and ENF~\citep{wessels2024equivariant}. For generation, we additionally compare against dedicated generative neural field based methods (GEM~\citep{du2021gem}, GASP~\citep{dupont2021gasp}, DPF~\citep{zhuang2023dpf}).

\vspace{-2mm}
\subsection{Reconstruction}
\label{sec:reconstruction}
\vspace{-1mm}

Table~\ref{tab:reconstruction} summarizes reconstruction quality across modalities. On both images and 3D shapes, LH-NeF consistently achieves state of the art performance among modality-agnostic neural field methods, while using significantly fewer parameters --up to $5{\times}$ less on CelebA $64^2$. On ImageNet1k~$256^2$, Spatial Functa reaches 38.4\,dB but uses a 2D conditioning latent grid of size $32{\times}32{\times}64{=}65,536$. At a comparable conditioning budget of size $8{\times}8{\times}256{=}16,384$ (vs.\ ours of $14,336$), we obtain competitive scores while providing a general modality-agnostic formulation (Spatial Functa is restricted to 2D signals). On ERA5, LH-NeF improves over Functa while remaining behind ENF, although we note that ENF relies on an equivariant formulation that needs prior knowledge of a symmetry group.

\vspace{-2mm}
\subsection{Downstream tasks}
\label{sec:downstream}
We evaluate whether the tokenizations learned by LH-NeF support downstream tasks they were never trained for, by training off-the-shelf models on the frozen tokenizations. Detailed experimental descriptions are provided in Appx.~\ref{app:downstream}.

\begin{wrapfigure}[18]{r}{0.43\textwidth}
\vspace{-10pt}
\centering
\begin{tikzpicture}
  % CIFAR-10 image
  \node[anchor=north east, inner sep=0] at (-0.05,0)
    {\includegraphics[width=0.37\textwidth]{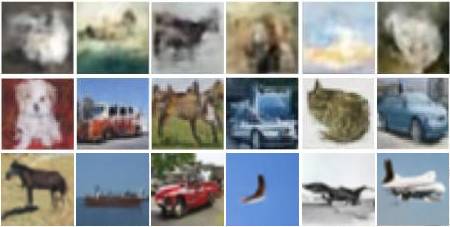}};
  % CIFAR-10 labels (x=-0.31\textwidth puts them left of image, adjust y to align rows)
  \node[anchor=east, font=\tiny\bfseries] at (-0.37\textwidth, -0.45) {Functa};
  \node[anchor=east, font=\tiny\bfseries] at (-0.37\textwidth, -1.35) {ENF};
  \node[anchor=east, font=\tiny\bfseries] at (-0.37\textwidth, -2.2) {\hspace{-.5em}LH-NeF};
  % CelebA-HQ image (y=-1.0 sets vertical gap from top image)
  \node[anchor=north east, inner sep=0] at (-0.05,-2.8)
    {\includegraphics[width=0.37\textwidth]{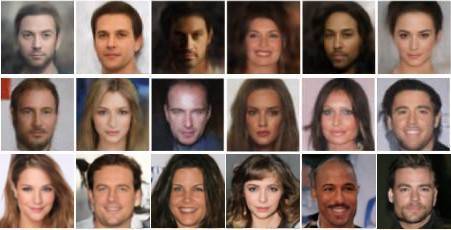}};
  % CelebA-HQ labels
  \node[anchor=east, font=\tiny\bfseries] at (-0.37\textwidth, -3.2) {Functa};
  \node[anchor=east, font=\tiny\bfseries] at (-0.37\textwidth, -4.1) {ENF};
  \node[anchor=east, font=\tiny\bfseries] at (-0.37\textwidth, -5) {\hspace{-.5em}LH-NeF};
\end{tikzpicture}
\captionsetup{font=small}
\caption{Unconditional generation on CIFAR-10 (top) and CelebA-HQ $64^2$ (bottom). Larger visualizations in Appx.~\ref{fig:celeba_samples_appx}.}
\label{fig:generation_samples}
\end{wrapfigure}
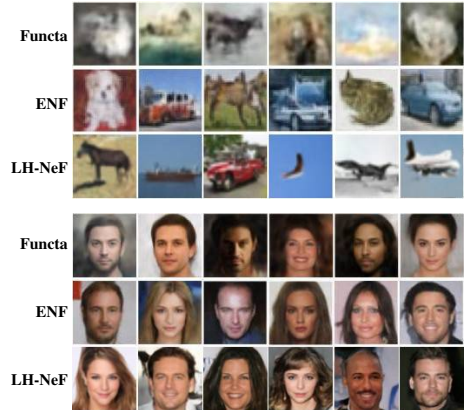
\textbf{Generation.}
We train a Diffusion Transformer \citep{peebles2023dit} on the (frozen) LH-NeF tokenizations.
On CelebA-HQ $64^2$, LH-NeF achieves state of the art generation (FID$\downarrow$), notably outperforming specialized generative methods and modality-agnostic neural field learning baselines (Table~\ref{tab:downstream}, samples in \cref{fig:generation_samples}).
On CIFAR-10, we achieve the best generation score among modality-agnostic representation learning methods, while remaining behind Spatial Functa and Diffusion Probabilistic Fields (DPF), which benefit from domain and task-specific advantages respectively.

\vspace{-2mm}
\paragraph{Classification.}
On ShapeNet16, a lightweight classifier on frozen latents reaches 96.8\% accuracy, outperforming ENF (96.6\%) and Functa (90.3\%).
On CIFAR-10, a ConvNeXt~\citep{liu2022convnet} classifier trained on the data's learned tokenizations reaches 91.2\% accuracy, outperforming all modality-agnostic baselines (ENF 82.1\%, Functa 68.3\%) as well as Spatial Functa~\citep{bauer2023spatial} (90.3\%, modality-specific).

\textbf{Global temperature forecasting.}
On ERA5, we follow the ENF protocol: consecutive hourly snapshots are tokenized, a temporal predictor maps $\mathbf{Y}_t {\mapsto} \mathbf{Y}_{t+1}$ in latent space, and predictions are decoded through the frozen renderer (details in Appx.~\ref{app:forecasting}).
LH-NeF improves over Functa but trails ENF, consistent with the reconstruction gap observed in Table~\ref{tab:reconstruction}; downstream predictions decode through the frozen renderer and are therefore upper-bounded by its fidelity.

\vspace{-2mm}
\subsection{Ablations and efficiency assessment}
\label{sec:ablations}
\vspace{-1mm}

\begin{table}[t]
\begin{minipage}[t]{0.66\textwidth}
\centering
\caption{Component ablation on CelebA-HQ $64^2$ (recon PSNR$\,\uparrow$, gen FID$\,\downarrow$) and ShapeNet16 (recon IoU$\,\uparrow$, cls acc$\,\uparrow$). ${}^\dagger$$k$-d tree on CelebA, Morton on ShapeNet.}
\label{tab:ablation}
\vspace{0.2em}
\scriptsize
\setlength{\tabcolsep}{2.5pt}
\renewcommand{\arraystretch}{1.0}
\resizebox{\textwidth}{!}{%
\begin{tabular}{@{}lcccc@{}}
\toprule
& \multicolumn{2}{c}{CelebA-HQ $64^2$} & \multicolumn{2}{c}{ShapeNet16} \\
\cmidrule(lr){2-3}\cmidrule(lr){4-5}
Configuration & PSNR $\uparrow$ & FID $\downarrow$ & IoU $\uparrow$ & Acc $\uparrow$ \\
\midrule
Full model & \textbf{36.3} & \textbf{9.7} & \textbf{93.2} & \textbf{96.8} \\
\midrule
\quad LH-NeF Tokenizer $\to$ HiP & 29.9 {\color{red}\tiny($-$6.4)} & 25.7 {\color{red}\tiny($+$16.0)} & 76.3 {\color{red}\tiny($-$16.9)} & 95.6 {\color{red}\tiny($-$1.2)} \\
\quad Other coord.\ ordering${}^\dagger$ & 35.5 {\color{red}\tiny($-$0.8)} & 10.7 {\color{red}\tiny($+$1.0)} & 92.5 {\color{red}\tiny($-$0.7)} & 96.0 {\color{red}\tiny($-$0.8)} \\
\quad Gaussian $\to$ power weight. & 34.1 {\color{red}\tiny($-$2.3)} & 13.4 {\color{red}\tiny($+$3.7)} & 85.5 {\color{red}\tiny($-$7.7)} & 96.8 {\tiny(\phantom{$+$}0.0)} \\
\quad $k \to 1$ (hard routing) & 35.6 {\color{red}\tiny($-$0.7)} & 10.2 {\color{red}\tiny($+$0.5)} & 84.8 {\color{red}\tiny($-$8.4)} & 96.7 {\color{red}\tiny($-$0.1)} \\
\quad No FiLM modulation & 34.0 {\color{red}\tiny($-$2.3)} & 13.0 {\color{red}\tiny($+$3.3)} & 88.9 {\color{red}\tiny($-$4.3)} & 97.0 {\tiny($+$0.2)} \\
\bottomrule
\end{tabular}}
\end{minipage}%
\hfill
\begin{minipage}[t]{0.32\textwidth}
\centering
\caption{Training efficiency on CelebA-HQ $64^2$ (H100 47\,GB MIG). Mem: peak step memory at $B{=}1$; Max BS: largest fitting batch.}
\label{tab:efficiency}
\vspace{0.2em}
\scriptsize
\setlength{\tabcolsep}{3pt}
\renewcommand{\arraystretch}{1.05}
\resizebox{\textwidth}{!}{%
\begin{tabular}{@{}lrrr@{}}
\toprule
 & Functa & ENF & \textbf{LH-NeF} \\
\midrule
Training step & MAML & MAML & Fwd.\,+\,Bwd. \\
Mem (MB) & 1{,}583 & 7{,}347 & \textbf{173} \\
Max BS & 18 & 2 & \textbf{266} \\
\bottomrule
\end{tabular}}
\end{minipage}
\vspace{-3mm}
\end{table}

Table~\ref{tab:ablation} ablates each component of LH-NeF on CelebA-HQ $64^2$ and ShapeNet16. Locality-preserving ordering is the most important component. Replacing it with raster ordering (vanilla HiP) produces ${-}6.4$\,dB PSNR and ${+}16.0$\,FID on CelebA; and ${-}16.9$\,IoU points on ShapeNet. Morton and $k$-d tree are near-interchangeable ($\leq 0.8$\,dB\,/\,IoU swap cost), confirming both locality-preserving orderings are effective. On the renderer side, Gaussian weighting and FiLM modulation are the most impactful for CelebA reconstruction (${-}2.3$\,dB each). Multi-group routing ($k{>}1$) contributes little on CelebA (${-}0.7$\,dB) but is the second-largest factor on ShapeNet (${-}8.4$\,IoU). Classification is largely unaffected by these changes, since it is a task that pools the spatial information and is therefore less affected by our locality ablation components. An extended discussion is provided in Appx.~\ref{app:ablation_analysis} including parallels with Spatial Functa~\citep{bauer2023spatial} and LIIF~\citep{chen2021liif}.

We benchmark the computational cost of LH-NeF according to the protocol described in Appx.~\ref{app:efficiency_protocol} (Table~\ref{tab:efficiency}). We use the official JAX implementations with XLA JIT compilation for Functa and ENF, which is favorable to the baselines as XLA fuses the MAML inner loop into a single optimized program. Even so, LH-NeF uses ${\sim}42{\times}$ less memory (173\,MB vs.\ 7.3\,GB) and supports $133{\times}$ larger batch sizes (266 vs.\ 2) than ENF on the same GPU. This results from second-order MAML methods requiring maintaining the full inner-loop computation graph across all $K$ inner steps, inflating per-sample memory by a factor of $K$ relative to a single forward-backward pass. Our method fits latents in one forward pass and still matches or improves over ENF in quality.

\vspace{-3mm}
\section{Conclusion}
\vspace{-2mm}

We presented LH-NeF, a framework for learning tokenizations of neural fields across metric coordinate domains that uses hierarchy and locality as inductive biases in its encoding and decoding process. LH-NeF achieves competitive or state-of-the-art performance on reconstruction and downstream tasks across 2D images, 3D shapes, and spherical climate data. By replacing per-sample meta-learning with a learned tokenization, our method uses significantly less memory and facilitates much larger batch sizes.
Altogether, LH-NeF serves as a scalable modality-agnostic foundational method for neural field representation learning.

\textbf{Limitations and broader perspective.}
\label{sec:limitations}
Fundamentally, our choice of leveraging hierarchy and locality biases is a \emph{deductive} approach motivated by structural properties observed in data across modalities.
A complementary, more \emph{inductive} approach is learning which biases generalize across tasks from a distribution over modalities and tasks~\citep{baxter2000bias}; this is a promising future work direction, as well as an active area of research, particularly on symmetry-related priors~\citep{romero2022partial,vanderlinden2024doublystochastic,urbano2026recon}. Generally, characterizing the full set of beneficial inductive biases across modalities, and understanding when learned biases should replace hand-designed ones, remains an open question. Another natural direction comes from defining a multi-hierarchy conditioning method that exploits information at all encoder levels rather than the last one, akin in spirit to multi-resolution hash encodings~\citep{mueller2022instant} or hierarchical latent diffusion~\citep{kim2023neuralfield_ldm}.
On signals with very irregular sampling (e.g.\ LiDAR), our locality guarantee holds only \emph{in expectation}: consecutive points in the locality-preserving order may fall into different groups, and soft routing in the renderer mitigates this only partially (see Appx.~\ref{app:ordering_tradeoffs}). A systematic study of the framework's behavior in that regime, along with scaling to substantially higher-resolution signals remains future work.

% Acknowledgements are handled via \begin{ack}...\end{ack} in neurips_2026.tex.

\begin{ack}
We thank David R. Wessels and David M. Knigge for their valuable discussions during the early stages of this research, and Carlos Saavedra Luque for his contributions to the design of the main figure of this paper.
This research was partially supported by the Deutsche Forschungsgemeinschaft (DFG) through the DFG Cluster of Excellence MATH+ (EXC-2046/1, EXC-2046/2, project id 390685689), as well as by the
German Federal Ministry of Research, Technology and Space (research campus Modal, fund number 05M14ZAM, 05M20ZBM) and the VDI/VDE Innovation + Technik GmbH (fund number 16IS23025B).
\end{ack}

\bibliographystyle{plainnat}
\bibliography{references}

%%%%%%%%%%%%%%%%%%%%%%%%%%%%%%%%%%%%%%%%%%%%%%%%%%%%%%%%%%%%
% APPENDIX
%%%%%%%%%%%%%%%%%%%%%%%%%%%%%%%%%%%%%%%%%%%%%%%%%%%%%%%%%%%%
\newpage
\appendix

%%%%%%%%%%%%%%%%%%%%%%%%%%%%%%%%%%%%%%%%%%%%%%%%%%%%%%%%%%%%%%%%%%%%%%%%%%%%%%%
%%  PART I — LOCALITY-PRESERVING ORDERINGS
%%%%%%%%%%%%%%%%%%%%%%%%%%%%%%%%%%%%%%%%%%%%%%%%%%%%%%%%%%%%%%%%%%%%%%%%%%%%%%%

\section{Coordinate Ordering Details}
\label{app:orderings}

\subsection{Euclidean Locality-preserving keys}
\label{app:orderings_euclidean}
\paragraph{Morton / Z-order key}
\label{app:morton}
Let $x\in[-1,1]^d$. Define the per-dimension integer quantization
$q_k(x) = \big\lfloor \tfrac{x^{(k)}+1}{2}\,(2^b-1) \big\rfloor \in \{0,\dots,2^b-1\}$,
where $b$ is the number of quantization bits.
Let $\mathrm{bit}_j(q_k(x))\in\{0,1\}$ denote the $j$-th least significant bit of $q_k(x)$.
The Morton key is obtained by interleaving bits across dimensions:
\begin{equation}
\kappa(x)=\sum_{j=0}^{b-1}\sum_{k=1}^{d} \mathrm{bit}_j(q_k(x))\,2^{dj+(k-1)}.
\end{equation}
Sorting tokens by $\kappa$ produces a space-filling curve that visits points in a spatially coherent order.
For $d{=}2$ this is the standard Z-curve; for $d{=}3$ it generalizes to a 3D Z-curve.
On regular grids, Morton ordering produces roughly square (2D) or cubic (3D) groups that respect spatial locality at every scale.

\paragraph{$k$-d tree linearization}
\label{app:kdtree}

For sparse or irregularly sampled data, Morton ordering can produce groups that span empty regions of the coordinate space (see \cref{app:ordering_tradeoffs}).
An alternative key is the $k$-d tree linearization:

\begin{enumerate}[leftmargin=*,itemsep=2pt]
\item Given a set of $N$ points with coordinates in $\mathbb{R}^d$, compute the bounding box and identify the axis of maximum spread.
\item Sort the points along that axis and split at the median.
\item Recurse on each half until each subset contains a single point.
\item Concatenate the leaves in depth-first, left-before-right order.
\end{enumerate}

The resulting permutation places spatially nearby points at adjacent positions in the sequence.

\subsection{Non-euclidean Locality-preserving keys}

\label{app:ordering_domains}

For data living on a non-Euclidean manifold, the same locality-preserving idea applies.
The construction is the same as in the Euclidean case (\cref{app:morton,app:kdtree}): pick any sort key whose level sets are spatially compact on the manifold, then use the resulting permutation as the input ordering for the contiguous-chunk grouping in the LH-NeF tokenizer.
\begin{figure}[h]
\centering
\small\textit{HiP}\par\vspace{4pt}
\hspace{2em}\includegraphics[width=0.28\linewidth]{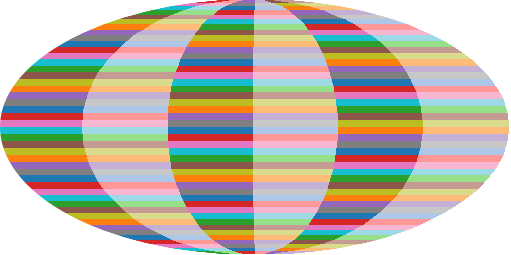}\hfill
\includegraphics[width=0.28\linewidth]{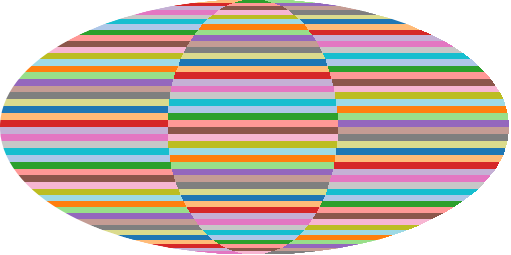}\hfill
\includegraphics[width=0.28\linewidth]{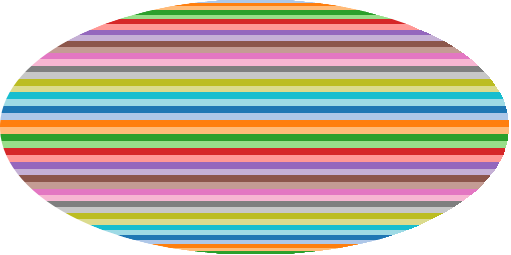}\hspace{2em}

\vspace{10pt}
\small\textit{LH-NeF (S2 Hilbert ordering)}\par\vspace{4pt}
\hspace{2em}\includegraphics[width=0.28\linewidth]{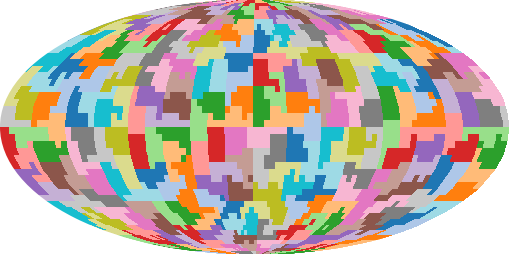}\hfill
\includegraphics[width=0.28\linewidth]{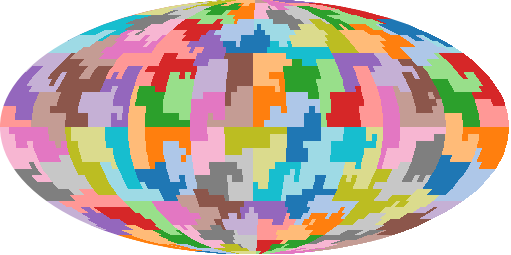}\hfill
\includegraphics[width=0.28\linewidth]{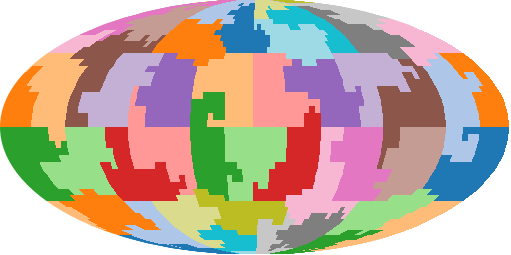}\hspace{2em}

\vspace{2pt}
{\hspace{4em}\small Block 1 ($G{=}288$) \hfill Block 2 ($G{=}144$) \hfill Block 3 ($G{=}48$)\hspace{4em}}
\caption{Receptive fields of the tokenizer hierarchy for the ERA5 ($46{\times}90$ lat-lon grid on $S^2$) LH-NeF tokenizer configuration.
\emph{HiP} (top): groups follow flat lat-lon raster order, producing latitude bands that lump unrelated climate regions together.
\emph{LH-NeF} (bottom): the S2 cell sort key produces compact, geographically coherent regions at every level.
Each color represents one group.}
\label{fig:hierarchy_era5}
\end{figure}

\paragraph{Spherical data (ERA5).}
For ERA5, points live on the unit sphere $S^2$ embedded as $(x,y,z) = (\cos\theta\cos\varphi,\,\cos\theta\sin\varphi,\,\sin\theta)$, which naturally handles periodicity at the antimeridian and convergence at the poles.
We sort points using the S2 cell hierarchy~\citep{s2geometry}: each point is mapped to its S2 cell token at a level chosen so that the total number of cells is at least $N$ (so that most cells contain at most one point), and the token strings are sorted lexicographically.
Because S2 cell IDs are constructed from a face index plus a Hilbert-curve position within that face, a lexicographic sort over tokens directly yields a Hilbert-curve traversal of the sphere.
We implement it using the \texttt{s2cell} Python package~\citep{s2cell-pypi}.

\subsection{Ordering on observations vs.\ coordinate space}
\label{app:ordering_tradeoffs}

Both orderings operate on the \emph{observed} coordinate-value tokens, not on a fixed discretization of the coordinate space.
Group boundaries are therefore determined by the distribution of observations. As a result, for uniformly sampled, regular data (e.g., images where every pixel is observed), the groups approximate a fixed spatial tiling and the induced regions are locality-preserving (compact) by construction.
For sparse or irregularly sampled data (e.g., point clouds), denser regions receive finer partitions while sparser regions receive coarser ones, adapting to the sampling density. This enables the same architecture to handle variable input sizes and sampling density without modification.

However, we note that on this case of non-uniformly sampled observations, locality-preserving holds only \emph{in expectancy}.
For example, on a 3D point cloud with very irregular sampling, consecutive points in Morton order may be assigned to a different group based on the total number of observations and number of groups used. While the use of $k$-soft group routing alleviates this partial locality violation, a systematic study of how this affects performance on highly irregularly sampled data such as LiDAR is left for future work.  

\subsection{Receptive field hierarchy}
\label{app:hierarchy}

\begin{figure}[h]
\centering
\small\textit{HiP}\par\vspace{4pt}
\hspace{2em}\includegraphics[width=0.16\linewidth]{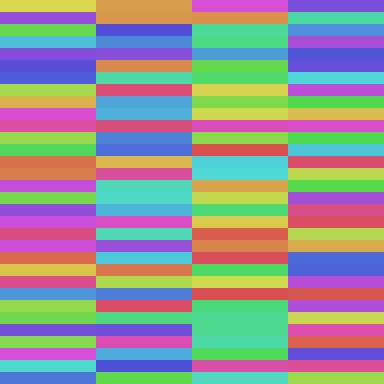}\hfill
\includegraphics[width=0.16\linewidth]{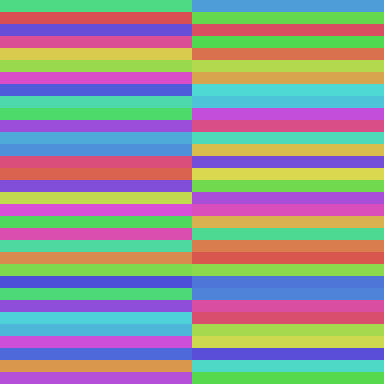}\hfill
\includegraphics[width=0.16\linewidth]{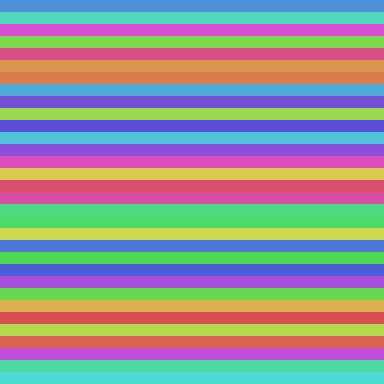}\hspace{2em}

\vspace{10pt}
\small\textit{LH-NeF (Morton ordering)}\par\vspace{4pt}
\hspace{2em}\includegraphics[width=0.16\linewidth]{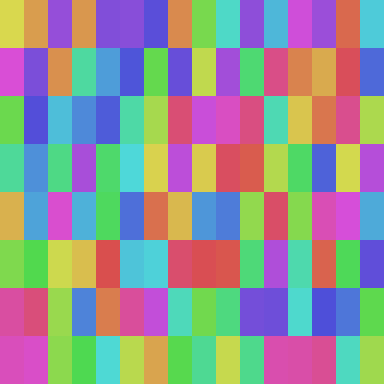}\hfill
\includegraphics[width=0.16\linewidth]{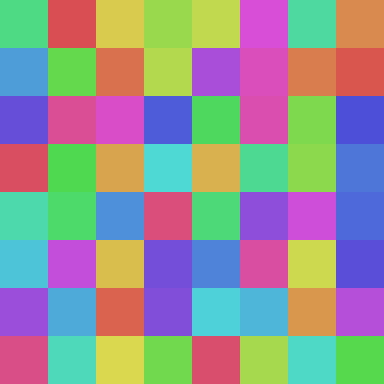}\hfill
\includegraphics[width=0.16\linewidth]{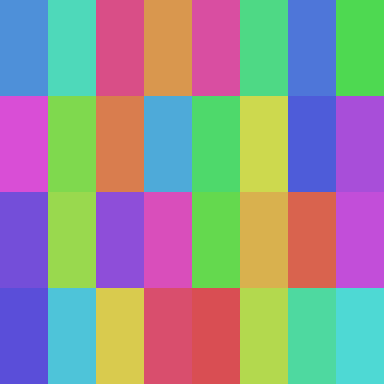}\hspace{2em}

\vspace{2pt}
{\hspace{2em}\small Block 1 ($G{=}128$) \hfill Block 2 ($G{=}64$) \hfill Block 3 ($G{=}32$)\hspace{2em}}
\caption{Receptive fields of the tokenizer hierarchy for the CIFAR-10 ($32{\times}32$) LH-NeF tokenizer configuration.
\emph{HiP} (top): groups are horizontal stripes with no 2D locality.
\emph{LH-NeF} (bottom): groups are spatially compact patches that coarsen hierarchically.
Each color represents one group.}
\label{fig:hierarchy_cifar10}
\end{figure}

\cref{fig:hierarchy_cifar10} visualizes the receptive fields induced by the LH-NeF tokenizer hierarchy on a $32{\times}32$ CIFAR-10 input.
Under HiP raster ordering (top row), each group covers a horizontal stripe, i.e. information from distant coordinates must be encoded in the same token group.
Under Morton ordering (bottom row), groups form compact, spatially coherent patches at every hierarchical level: from $128$ fine patches at block~1, through $64$ medium patches at block~2, to $32$ coarse regions at block~3.
Note that groups at blocks~1 and~3 appear as $2{:}1$ rectangles rather than perfect squares: this is because the Z-curve alternates which spatial dimension it doubles at each recursion level, so contiguous chunks of $2^{2m}$ tokens form squares while chunks of $2^{2m+1}$ tokens form $2{:}1$ rectangles. Groups remain spatially compact however, and soft group routing aggregates group information smoothly regardless.

\begin{figure}[h]
\centering
\begin{tabular}{@{}c@{\hskip 4pt}c@{\hskip 4pt}c@{\hskip 4pt}c@{}}
\includegraphics[width=0.2\linewidth]{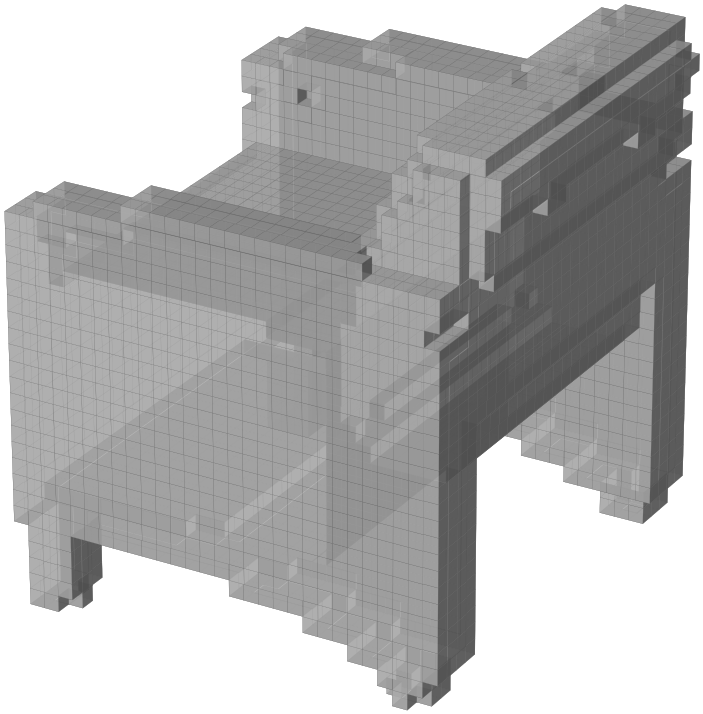} &
\hspace{1.75em}\includegraphics[width=0.2\linewidth]{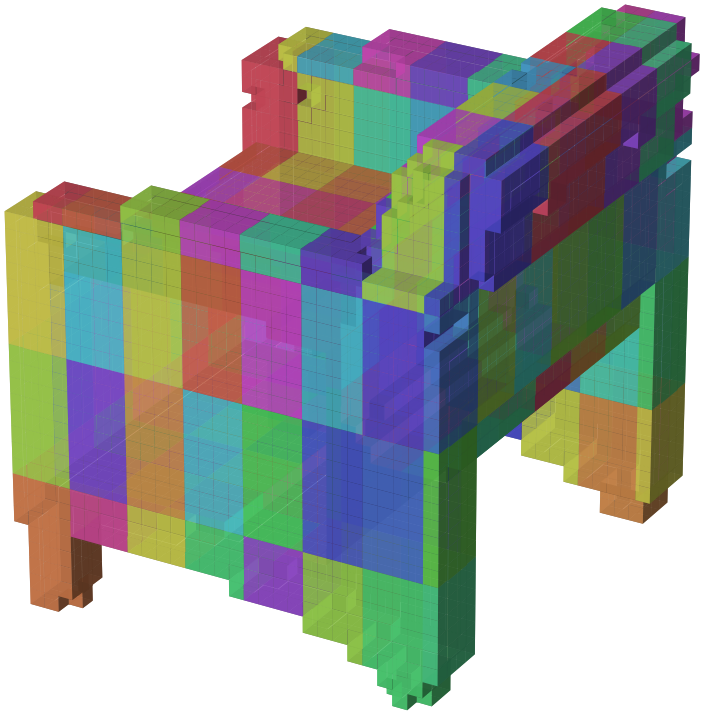} &
\hspace{1.75em}\includegraphics[width=0.2\linewidth]{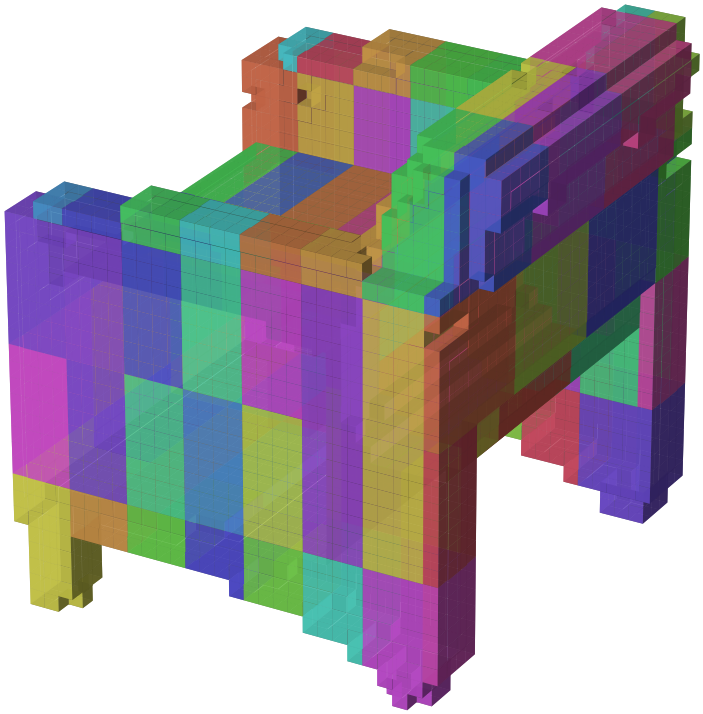} &
\includegraphics[width=0.2\linewidth]{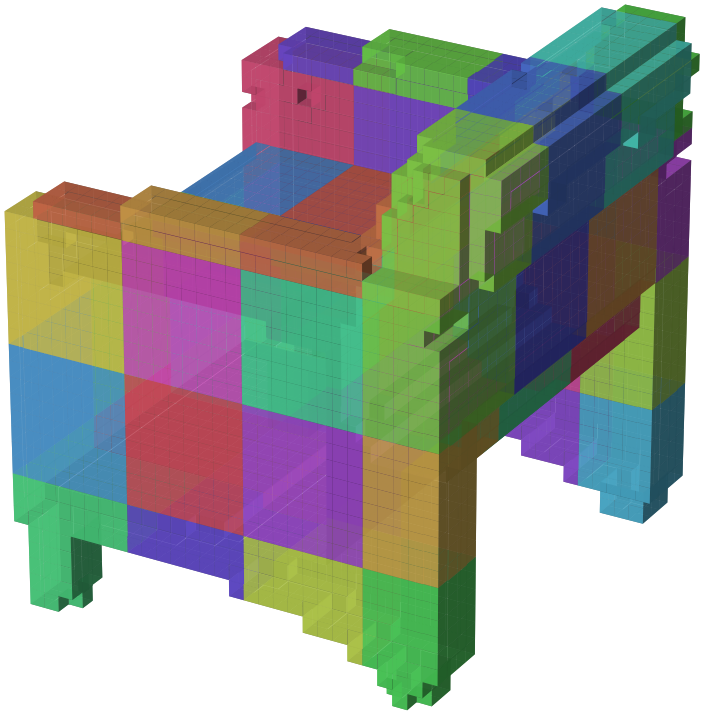} \\[1pt]
{\small Input} & {\small Block 1 ($G{=}256$)} & {\small Block 2 ($G{=}128$)} & {\small Block 3 ($G{=}64$)} \\
\end{tabular}
\caption{Receptive field LH-NeF tokenizer hierarchy on a ShapeNet chair ($32^3$ voxels, $k$-d tree ordering).}
\label{fig:hierarchy_shapenet}
\end{figure}

\cref{fig:hierarchy_shapenet} shows the same visualization on a ShapeNet chair ($32^3$ voxels) using $k$-d tree ordering.
Since grouping depends on the observed voxel distribution, groups adapt to the shape geometry, i.e. different inputs get different group distributions (and therefore routing metadata).

\cref{fig:hierarchy_era5} shows the same visualization for ERA5 global temperature ($46{\times}90$ lat-lon grid on the unit sphere $S^2$) under our LH-NeF tokenizer with the S2 cell ordering described in \cref{app:ordering_domains}.
Under flat lat-lon raster ordering each group is a horizontal latitude band that spans the entire globe; under S2 ordering, groups become compact, geographically coherent patches at every level of the hierarchy.

\paragraph{Spatial partition on $\mathcal{X}$ induced by the LH-NeF tokenizer}
\label{app:voronoi}

\begin{figure}[h]
\centering
\includegraphics[width=0.20\linewidth]{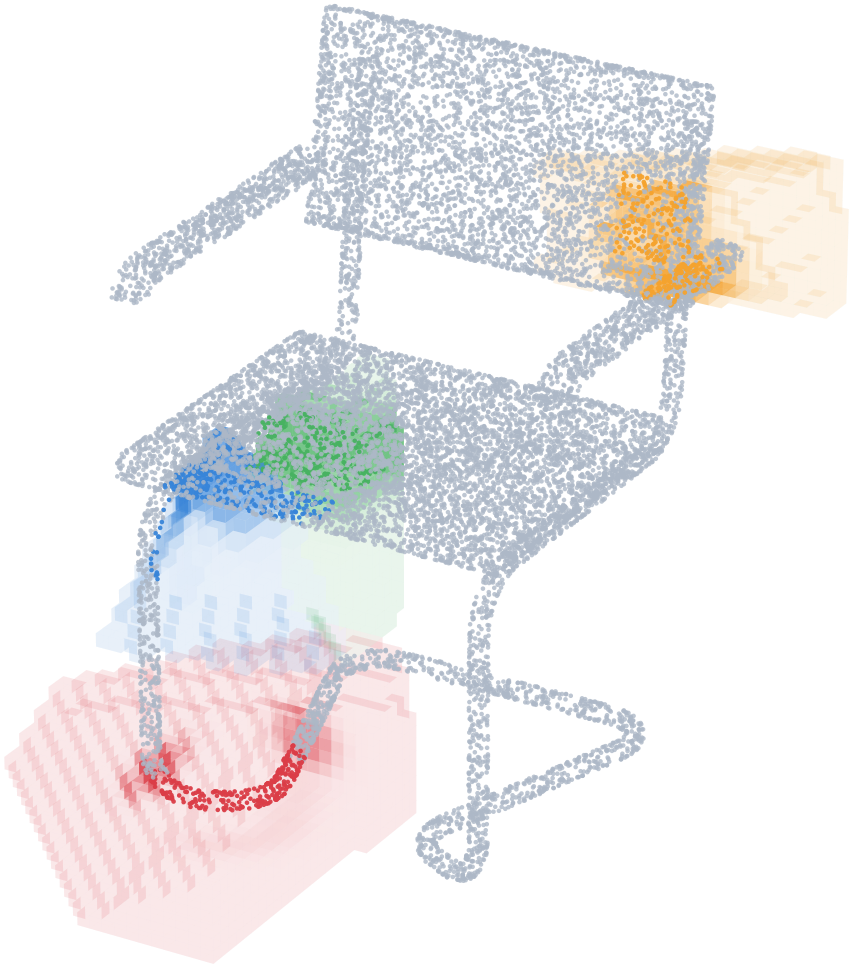}\hfill
\includegraphics[width=0.20\linewidth]{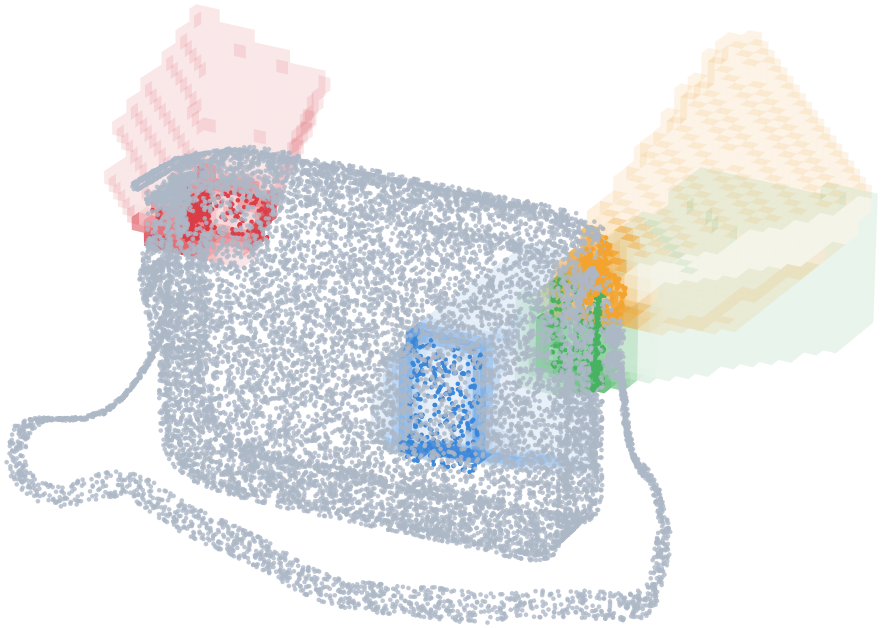}\hfill
\includegraphics[width=0.21\linewidth]{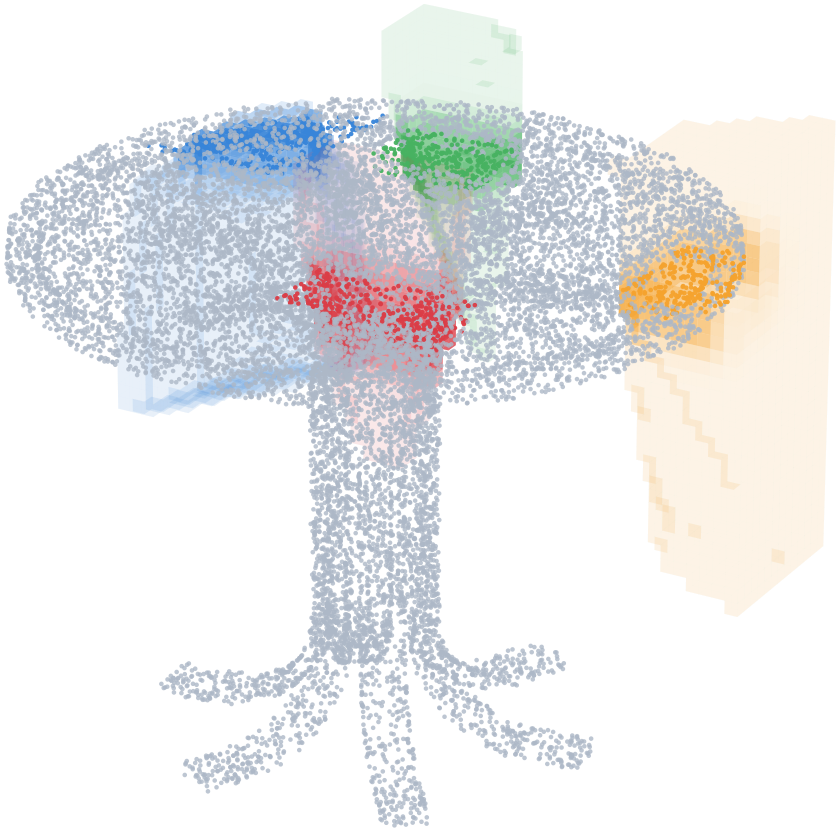}\hfill
\includegraphics[width=0.13\linewidth]{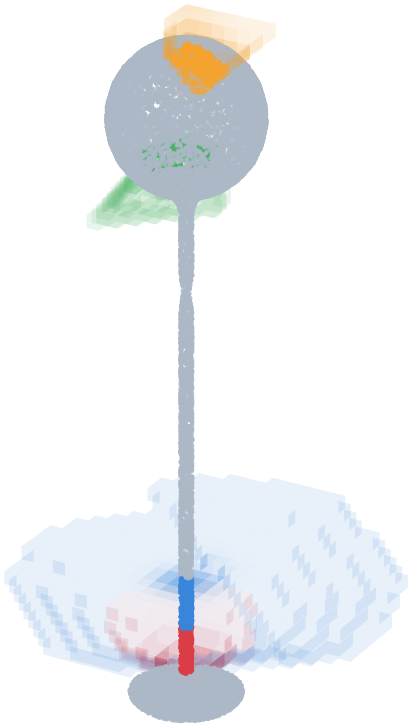}
\caption{Voronoi partition of 3D space induced by $k$-d tree group centroids ($G{=}64$, four groups highlighted per shape).
Colored points: observed surface samples belonging to each highlighted group.
Semi-transparent colored cells: the corresponding Voronoi regions extending into unoccupied space. Groups adapt each input's point cloud geometry.
From left to right: chair, bag, table, lamp.}
\label{fig:voronoi_3d}
\end{figure}

The LH-NeF renderer assigns each query coordinate to its $k$-nearest group centroids.
Note that this mechanism induces a partition on the \emph{entire} continuous coordinate space, not just the observed points. This matters on e.g. point clouds. We visualize these partitions on different ShapeNet point clouds, illustrating how (arbitrary) query coordinates in unoccupied, distant regions would be routed to nearby surface groups (\cref{fig:voronoi_3d}). 
Colored points show observations belonging to each highlighted group, while the semi-transparent colored cells show the full 3D Voronoi-like region that would assign queries to that group.
Cells on the object surface extend into the surrounding empty space: any query in that region is routed to the nearest surface group(s).
Each shape induces a different partition since group centroids depend on the distribution of observed samples.

\begin{figure}[h]
\centering
\includegraphics[width=.9\textwidth]{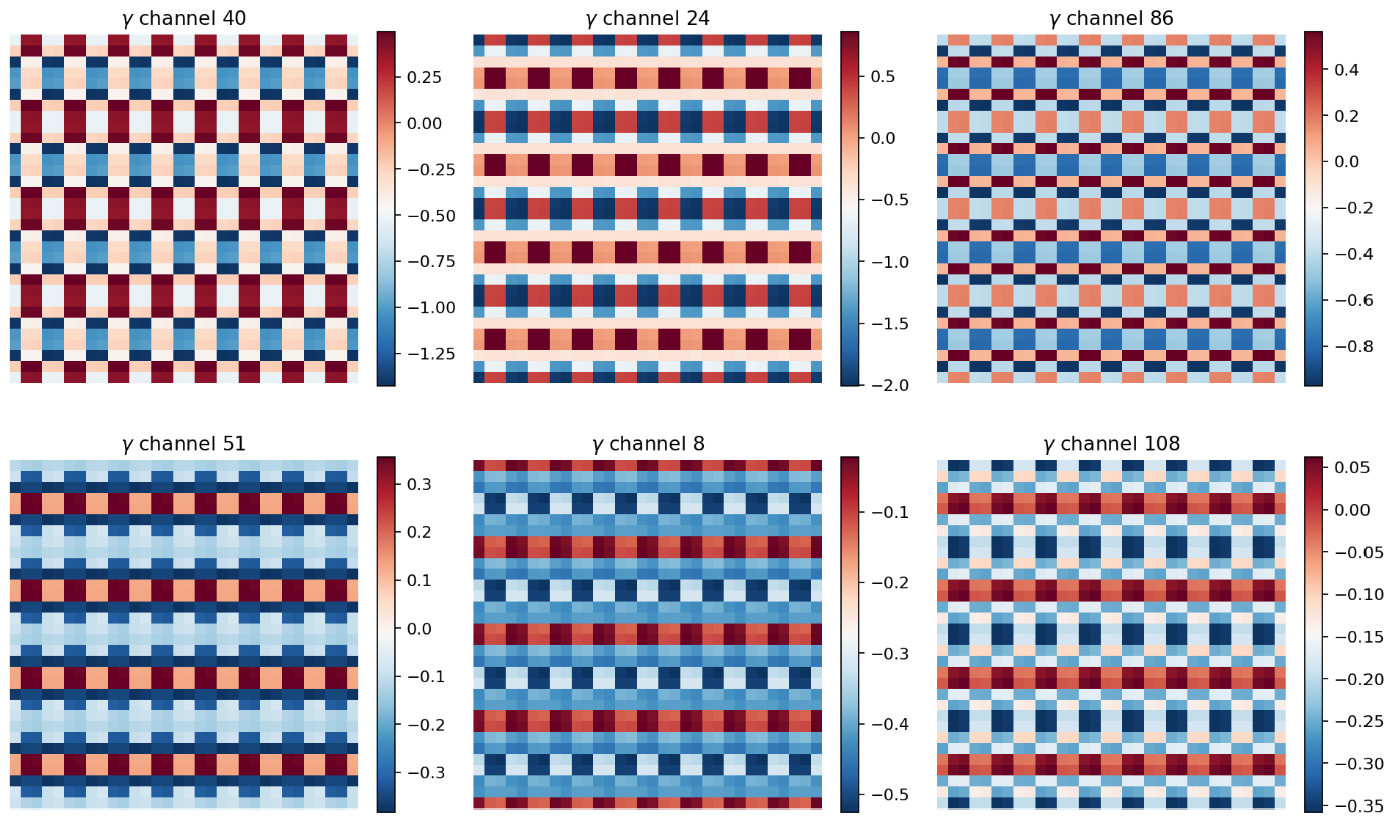}
\caption{FiLM scale parameter $\gamma$ (six different channels) evaluated on a $32{\times}32$ query grid with the trained CIFAR10 checkpoint.
The modulation pattern repeats identically across all spatial groups (see overlay in \cref{fig:hierarchy_cifar10} Block 3),
showing how the group-normalized coordinate frame
$\tilde{x} = (x - \mu_g) / \lambda_g$ makes the FiLM output
invariant to the spatial position of the group.
Since all groups share the same spatial extent on uniform grids,
this figure demonstrates translation invariance;
on variable-geometry data (e.g.\ point clouds), the same property
additionally ensures scale invariance across groups of different sizes.}
\label{fig:film_tiling}
\end{figure}
 
\subsection{Invariance of the FiLM coordinate frame}
\label{app:film_invariance}

The FiLM modulation in the LH-NeF renderer (\cref{sec:renderer})
conditions on the group-normalized relative coordinate $\tilde{x}$.
We show that $\tilde{x}$ is invariant under natural families of
transformations, so that the FiLM output
$(\gamma,\beta) = \mathrm{MLP}_{\mathrm{FiLM}}(\mathrm{PE}(\tilde{x}))$
is identical for the same local pattern at any absolute position or scale.

\paragraph{Euclidean domains.}
On $\mathcal{X} \subseteq \mathbb{R}^d$, the group-normalized coordinate
is $\tilde{x}^{\,g} = (x - \mu_g) / \lambda_g$ with
$\mu_g = \frac{1}{|\mathcal{I}_g|}\sum_i x_i$ and
$\lambda_g = \max_i x_i - \min_i x_i \in \mathbb{R}^d_{>0}$
(\cref{sec:encoder}).
Under any transformation $T_{t,S}: x \mapsto Sx + t$ with
$t \in \mathbb{R}^d$ and
$S = \mathrm{diag}(s_1,\ldots,s_d)$, $s_k > 0$,
applied jointly to all observations and the query,\footnote{For
translations and uniform scalings the transformation is a similarity
and preserves the nearest-group assignment
$g^* = \argmin_g d_{\mathcal{X}}(x, \mu_g)$. For anisotropic
per-axis scalings ($S$ with non-equal diagonal entries), routing
preservation is not guaranteed in general, since stretching one axis
can swap which centroid is nearest.}
the centroid transforms as $\mu'_g = S\mu_g + t$ and the extent as
$\lambda'_g = S \lambda_g$ (element-wise, since $S$ is a positive
diagonal), giving
\begin{equation}
\tilde{x}'
  = \frac{(Sx+t) - (S\mu_g+t)}{S\lambda_g}
  = \frac{S(x-\mu_g)}{S\lambda_g}
  = \frac{x-\mu_g}{\lambda_g}
  = \tilde{x}.
\end{equation}
This covers translations ($S=I$), uniform scaling ($S=sI$), and
anisotropic per-axis rescaling (distinct $s_k$). The per-axis
bounding-box extent is what enables the anisotropic case: each
coordinate dimension normalizes independently, so the diagonal
entries of $S$ cancel component-wise.

\paragraph{General Riemannian manifolds.}
On a complete Riemannian manifold $(\mathcal{X}, d_{\mathcal{X}})$
with scalar group radius
$\lambda_g = \max_i d_{\mathcal{X}}(\mu_g, x_i)$, the analogous result holds for geodesic
dilations $D_s(x) = \exp_{\mu_g}(s \cdot \log_{\mu_g}(x))$,
$s > 0$.
Let $v_i = \log_{\mu_g}(x_i)$ and $v = \log_{\mu_g}(x)$.
The Fréchet mean is preserved because $D_s$ scales all tangent
vectors by $s$, so the optimality condition
$\sum_i v_i = 0$ becomes $s\sum_i v_i = 0$, which still holds
under the same uniqueness conditions~\citep{afsari2011riemannian}.
The radius scales to $s\lambda_g$ (since
$d_{\mathcal{X}}(\mu_g, \exp_{\mu_g}(sv_i)) = s\|v_i\|_{\mu_g}$),
and $\log_{\mu_g}(D_s(x)) = sv$, giving
\begin{equation}
\tilde{x}'
  = \frac{sv}{s\lambda_g}
  = \frac{v}{\lambda_g}
  = \tilde{x}.
\end{equation}
On $\mathbb{R}^d$, geodesic dilation reduces to
$x \mapsto \mu_g + s(x-\mu_g)$, i.e.\ the uniform-scaling
subcase $S = sI$ of the Euclidean result above.
The Euclidean formulation is strictly stronger: translations and
anisotropic rescalings are additionally available because the
per-axis bounding-box extent normalizes each coordinate dimension
independently (structure that is unavailable on a general manifold).

The invariance provides an inductive bias analogous to weight sharing in convolutional networks: the FiLM MLP learns a single mapping from relative position to modulation parameters, and the normalization by $(\mu_g, \lambda_g)$ ensures this mapping generalizes across groups at different absolute locations and of different spatial extents without requiring the network to learn position- or scale-dependent features.
The ablation in Table~\ref{tab:ablation} confirms the empirical impact: removing FiLM modulation costs $2.3$\,dB PSNR on CelebA-HQ and $4.3$ IoU on ShapeNet.

%%%%%%%%%%%%%%%%%%%%%%%%%%%%%%%%%%%%%%%%%%%%%%%%%%%%%%%%%%%%%%%%%%%%%%%%%%%%%%%
%%  PART II — STAGE 1: RECONSTRUCTION
%%%%%%%%%%%%%%%%%%%%%%%%%%%%%%%%%%%%%%%%%%%%%%%%%%%%%%%%%%%%%%%%%%%%%%%%%%%%%%%

\section{LH-NeF Implementation Details}
\label{app:stage1}

\subsection{LH-NeF tokenizer configuration}
\label{app:encoder}

The LH-NeF tokenizer processes re-ordered inputs with Hierarchical Perceiver~\citep{rabe2021selfattention} grouped attention blocks.
Each block is characterized by its number of groups $G$, tokens per group $K$, channels $C$, and number of self-attention layers.
Tokens are progressively regrouped from many small groups (fine locality) to fewer, larger groups (coarser, more global context).
The conditioning representation is the output of the final block: a structured token set of shape $[B, G, K, C]$ that serves as the per-sample representation for the LH-NeF renderer.
Unlike the original HiP, our bottleneck representation is multi-group (rather than single group), since we aim to exploit the locality of each group at query time.

Table~\ref{tab:encoder_hierarchy} shows the block hierarchy for each dataset.
Compute is concentrated at the final block (18-36 self-attention layers) with lighter processing at earlier stages (1-3 layers).

\begin{table}[h]
\centering
\caption{LH-NeF tokenizer block hierarchy per dataset.
Each row shows the LH-NeF tokenizer variant used on each dataset; the conditioning representation is extracted from the final block.
$G$: groups, $K$: tokens/group, $C$: channels, SA: self-attention layers. ImageNet uses 4 blocks (block 0--3); all others use 3 (block 0--2).}
\label{tab:encoder_hierarchy}
\small
\setlength{\tabcolsep}{3.5pt}
\begin{tabular}{l ccc ccc ccc ccc c}
\toprule
& \multicolumn{3}{c}{Block 0} & \multicolumn{3}{c}{Block 1} & \multicolumn{3}{c}{Block 2} & \multicolumn{3}{c}{Block 3} &  \\
\cmidrule(lr){2-4} \cmidrule(lr){5-7} \cmidrule(lr){8-10} \cmidrule(lr){11-13}
Dataset & $G$ & $K$/$C$ & SA & $G$ & $K$/$C$ & SA & $G$ & $K$/$C$ & SA & $G$ & $K$/$C$ & SA & Cond. dims \\
\midrule
CIFAR-10      & 128 & 4/64 & 1 & 64 & 4/96  & 2  & 32 & 4/16  & 18 & -- & -- & -- & 2,048 \\
CelebA-HQ $64^2$ & 256 & 4/96 & 1 & 128 & 4/128 & 3  & 64 & 2/24  & 36 & -- & -- & -- & 3,072 \\
ImageNet1k $256^2$ & 1,024 & 4/64 & 1 & 512 & 4/96 & 2 & 256 & 4/128 & 24 & 128 & 4/28 & 6 & 14,336 \\
ShapeNet16    & 256 & 4/64 & 1 & 128 & 4/32  & 2  & 64 & 3/8   & 24 & -- & -- & -- & 1,536 \\
ERA5          & 288 & 4/96 & 1 & 144 & 4/128 & 3  & 48 & 2/24  & 36 & -- & -- & -- & 2,304 \\
\bottomrule
\end{tabular}
\end{table}

\paragraph{Input embedding.}
All variants use sinusoidal positional encoding on input coordinates before projection into the HiP embedding space.
Table~\ref{tab:encoder_embedding} lists the embedding hyperparameters.

\begin{table}[h]
\centering
\caption{LH-NeF input embedding and ordering configuration per dataset.}
\label{tab:encoder_embedding}
\small
\begin{tabular}{lccccc}
\toprule
& CIFAR-10 & CelebA-HQ $64^2$ & ImageNet1k $256^2$ & ShapeNet16 & ERA5 \\
\midrule
Coord.\ dim $d$ & 2 & 2 & 2 & 3 & 3 \\
Value dim & 3 (RGB) & 3 (RGB) & 3 (RGB) & 1 (occ.) & 1 (temp.) \\
Fourier bands & 16 & 32 & 32 & 32 & 32 \\
Max frequency & 40 & 40 & 40 & 20 & 20 \\
Embedding channels & 64 & 64 & 64 & 64 & 128 \\
Coord.\ ordering & Morton & Morton & Morton & $k$-d tree & S2 cell \\
\bottomrule
\end{tabular}
\end{table}

\subsection{LH-NeF renderer configuration}
\label{app:renderer}

Table~\ref{tab:renderer_hparams} lists the LH-NeF renderer hyperparameters per dataset.
All configurations render from the LH-NeF tokenization, with a point MLP head,
Gaussian-windowed KNN routing with learnable $\sigma$, and FiLM modulation.

\begin{table}[h]
\centering
\caption{LH-NeF renderer hyperparameters per dataset.}
\label{tab:renderer_hparams}
\small
\begin{tabular}{lccccc}
\toprule
& CIFAR-10 & CelebA-HQ $64^2$ & ImageNet1k $256^2$ & ShapeNet16 & ERA5 \\
\midrule
$d_\text{model}$ (cross-attn) & 128 & 128 & 256 & 256 & 128 \\
Num.\ heads & 8 & 8 & 8 & 8 & 4 \\
$k$ (KNN groups) & 4 & 4 & 4 & 8 & 2 \\
$\sigma_\text{init}$ (Gaussian) & 0.25 & 0.35 & 0.5 & 0.75 & 0.35 \\
MLP hidden & 128 & 384 & 768 & 256 & 192 \\
MLP depth & 3 & 5 & 4 & 4 & 3 \\
FiLM hidden & 256 & 256 & 256 & 256 & 512 \\
FiLM PE dim & 16 & 16 & 16 & 8 & 8 \\
Coord PE dim & 32 & 16 & 16 & 32 & 32 \\
\bottomrule
\end{tabular}
\end{table}

\paragraph{Conditioning dimensions.}
The $G \times K \times C$ token representation is the primary per-sample conditioning.
In addition, the LH-NeF renderer receives per-group \emph{routing metadata}: the group center $\mu_g \in \mathbb{R}^d$ (mean coordinate of input points assigned to group $g$) and group scale $\lambda_g \in \mathbb{R}^d$ (bounding-box extent), for a total of $2Gd$ additional dimensions.
These are computed from the input \emph{coordinates} only (i.e. they are not learnable parameters); the renderer uses them for KNN group routing and relative-coordinate normalization in the FiLM path.
For fixed-grid data (images, ERA5, voxels), $\mu_g$ and $\lambda_g$ are constant across samples and carry zero bits of per-sample information.
For variable-input data (e.g. point clouds), the routing metadata varies per sample.
Table~\ref{tab:full_budget} reports all dimensions per dataset.

\begin{table}[h]
\centering
\caption{Per-sample representation dimensions: token representation ($G \times K \times C$) and routing metadata ($2Gd$ dimensions for group centers and scales).}
\label{tab:full_budget}
\small
\begin{tabular}{lcccc}
\toprule
Dataset & $d$ & Tokens dim.\ ($GKC$) & Routing dim.\ ($2Gd$) & Total \\
\midrule
CIFAR-10       & 2 & 2,048 & 128$^\ast$  & 2,176 \\
CelebA-HQ $64^2$ & 2 & 3,072 & 256$^\ast$ & 3,328 \\
ImageNet1k $256^2$ & 2 & 14,336 & 512$^\ast$ & 14,848 \\
ShapeNet16     & 3 & 1,536 & 384$^\ast$         & 1,920 \\
ERA5           & 3 & 2,304 & 288$^\ast$  & 2,592 \\
\bottomrule
\end{tabular}
\\[3pt]
{\scriptsize $^\ast$\,Constant across samples (fixed-grid, uniformly sampled data).}
\end{table}

\subsection{Training configuration}
\label{app:training}

All datasets use AdamW for the tokenizer and Adam for the renderer, with cosine learning rate decay.
Checkpoints are selected by the primary reconstruction metric: PSNR for images, IoU for 3D occupancy, MSE for ERA5.

\begin{table}[h]
\centering
\caption{LH-NeF training configuration per dataset.}
\label{tab:training_details}
\small
\begin{tabular}{lccccc}
\toprule
& CIFAR-10 & CelebA-HQ $64^2$ & ImageNet1k $256^2$ & ShapeNet16 & ERA5 \\
\midrule
Tokenizer observations $N$ & 1,024 & 4,096 & 65,536 & 32,768 & 4,140 \\
Batch size & 384 & 128 & 52 & 40 & 64 \\
Epochs & 1,000 & 1,000 & 100 & 300 & 1,000 \\
Tokenizer LR & 1.2e-3 & 2.8e-4 & 5.5e-4 & 1.3e-3 & 1.8e-3 \\
Tokenizer weight decay & 5.3e-3 & 1.6e-4 & 1.7e-2 & 2.1e-3 & 7.0e-3 \\
Renderer LR & 1.8e-4 & 9.0e-4 & 1.6e-4 & 4.9e-4 & 2.7e-4 \\
Loss & L1 & L1 & L1 & L1 & MSE \\
Selection metric & PSNR $\uparrow$ & PSNR $\uparrow$ & PSNR $\uparrow$ & IoU $\uparrow$ & MSE $\downarrow$ \\
GPUs & 1$\times$A40 & 1$\times$A100 & 2$\times$A100 & 2$\times$H200 & 1$\times$A40 \\
Training time & ${\sim}$18\,h & ${\sim}$29\,h & ${\sim}$87\,h & ${\sim}$25\,h & ${\sim}$64\,h \\
\bottomrule
\end{tabular}
\end{table}

\section{Detailed Component Ablation Analysis}
\label{app:ablation_analysis}

We extend the discussion of Table~\ref{tab:ablation} from the main text.

\paragraph{Locality-preserving ordering.}
Locality dominates the ablations on dense tasks (reconstruction, generation), but classification is the exception ($-1.2$\,pp without locality). This is because our classifier pools the spatial structure of the LH-NeF tokens via a global \texttt{[CLS]} readout (Appx.~\ref{app:classification}), which discards geometric structure from the representation. Dense tasks like reconstruction and generation rely directly on the spatial structure of the tokens, where locality has a much larger effect.

\paragraph{Renderer components on CelebA-HQ reconstruction.}
\emph{Gaussian weighting} and \emph{relative-coordinate FiLM modulation} tie at $-2.3$\,dB each, while \emph{multi-group routing} ($k{>}1$) contributes less ($-0.7$\,dB). The latter result mirrors Spatial Functa~\citep{bauer2023spatial}, which reports that single-neighbor latent lookup matches or beats bilinear interpolation on images, and LIIF~\citep{chen2021liif}, which reports small but consistent gains from using four local neighbors for super-resolution -- a task in which the field is queried often close to group boundaries, where continuity through multiple neighbors is more important.

\paragraph{Renderer components on CelebA-HQ generation.}
For generation, every renderer component except $k{>}1$ becomes significant, and both locality-preserving orderings remain effective with minor gains for Morton over $k$-d tree.

\paragraph{Renderer components on ShapeNet16.}
The ranking flips on ShapeNet: $k{>}1$ and Gaussian weighting become the second most impactful ($-8.4$ and $-7.7$\,IoU). Multiple neighbors help on ShapeNet because the renderer queries a full $32^3$ grid that includes many empty-space voxels: queries near shape boundaries can borrow representational capacity from empty-space group tokens via multi-group routing. Renderer ablations barely move ShapeNet classification ($<$1\,pp), by the same global-readout argument as above.

%%%%%%%%%%%%%%%%%%%%%%%%%%%%%%%%%%%%%%%%%%%%%%%%%%%%%%%%%%%%%%%%%%%%%%%%%%%%%%%
%%  PART III — DOWNSTREAM TASKS
%%%%%%%%%%%%%%%%%%%%%%%%%%%%%%%%%%%%%%%%%%%%%%%%%%%%%%%%%%%%%%%%%%%%%%%%%%%%%%%

\section{Downstream Task Implementation Details}
\label{app:downstream}

All downstream models are trained on frozen tokenizations with no gradient flow to the tokenizer or renderer.

\subsection{Generation (Latent Diffusion)}
\label{app:generation}

Generation uses a diffusion transformer trained on the (frozen) tokenizations learned during the LH-NeF training (reconstruction stage).
Training and evaluation follows three steps: (i) preprocessing (tokenization extraction and normalization), (ii) diffusion model training, and (iii) sampling and rendering for FID computation.

\paragraph{Tokenization extraction and normalization.}
Given a frozen LH-NeF model, we run the LH-NeF tokenizer on every training and validation sample, extracting the grouped token representation $\mathbf{Y}^{(L)} \in \mathbb{R}^{G_L \times K_L \times C_L}$ from the final block (for brevity we write $G, K, C$ for $G_L, K_L, C_L$ throughout this section).
Along with each token set we store the group routing metadata (centroids $\mu_g \in \mathbb{R}^d$ and extents $\lambda_g \in \mathbb{R}^d_{>0}$).
Tokenizations are cast to float16 and written to sharded archives for efficient loading.
We compute running per-channel statistics (mean, std) across all training samples; during diffusion training, tokenizations are normalized as $\hat{\mathbf{Y}}^{(L)} = (\mathbf{Y}^{(L)} - \mathbf{m}) / \mathbf{s}$ where $\mathbf{m}, \mathbf{s} \in \mathbb{R}^C$ are the per-channel mean and standard deviation (broadcast over groups and tokens).

\paragraph{Diffusion model architecture.}
The target of the latent diffusion model is the (normalized) tokenization $\hat{\mathbf{Y}}^{(L)} \in \mathbb{R}^{G \times K \times C}$. We use a DiT variant adapted to the grouped structure: self-attention within each group's $K$ tokens, and regrouping across blocks via cross-attention from learnable latent queries ($G \to G/4 \to 1$), so that a single global-attention layer at the coarsest level provides full cross-group communication. A symmetric decoder path with skip connections expands back to the original grouping (U-Net style). The grouping reduces self-attention cost from $\mathcal{O}((GK)^2)$ to $\mathcal{O}(GK^2 + G^2)$ across the hierarchy.

Following DiT~\citep{peebles2023dit}, all sublayers use adaLN-Zero conditioning from a noise-level embedding (the EDM $c_\text{noise}(\sigma)$ passed through random Fourier features), with zero-initialized gates. Continuous coordinate embeddings are computed from the group centroids $\mu_g$ via a random Fourier feature projection. Since all $K$ tokens within a group share the same positional embedding (their group centroid $\mu_g$), we add a learned per-position token-ID embedding ($j \in \{1, \ldots, K\}$) to break within-group permutation symmetry.

\paragraph{Training.}
We adopt the EDM formulation~\citep{karras2022edm}. Writing $\mathbf{z}_0 = \hat{\mathbf{Y}}^{(L)}$ for the clean (normalized) tokenization, we perturb it with continuous, variance-exploding noise $\mathbf{z}_\sigma = \mathbf{z}_0 + \sigma\,\epsilon$, and wrap the network with Karras preconditioning so that the denoised estimate is
\begin{equation}
\hat{\mathbf{z}}_0 = c_\text{skip}(\sigma)\,\mathbf{z}_\sigma + c_\text{out}(\sigma)\,F_\theta\big(c_\text{in}(\sigma)\,\mathbf{z}_\sigma,\, c_\text{noise}(\sigma)\big), \quad c_\text{noise}(\sigma) = \tfrac{1}{4}\ln\sigma,
\end{equation}
with the standard $c_\text{skip}, c_\text{out}, c_\text{in}$ coefficients of~\citet{karras2022edm}.
Noise levels are sampled as $\ln\sigma \sim \mathcal{N}(P_\text{mean}, P_\text{std}^2)$ ($P_\text{mean}{=}{-}1.2$, $P_\text{std}{=}1.2$), and we minimize the EDM-weighted denoising loss
\begin{equation}
\mathcal{L}_\text{dm} = \mathbb{E}_{\sigma, \epsilon}\big[\lambda(\sigma)\,\| \hat{\mathbf{z}}_0 - \mathbf{z}_0 \|^2\big], \quad
\lambda(\sigma) = \frac{\sigma^2 + \sigma_\text{data}^2}{(\sigma\,\sigma_\text{data})^2},
\end{equation}
where $\sigma_\text{data}$ is estimated from the normalized tokenizations ($\approx 1$).
We use AdamW and maintain an exponential moving average (EMA) of model weights for sampling.

\paragraph{Sampling.}
We generate samples with the deterministic ODE sampler of~\citet{karras2022edm}, using 18 steps over the Karras noise schedule ($\rho{=}7$, $\sigma_\text{min}{=}0.002$, $\sigma_\text{max}{=}80$, no stochastic churn).
Sampled tokenizations are denormalized ($\mathbf{Y}^{(L)} = \hat{\mathbf{Y}}^{(L)} \cdot \mathbf{s} + \mathbf{m}$), reshaped to $[B, G, K, C]$, and passed through the frozen LH-NeF renderer to produce images.
The renderer receives the sampled tokens as its tokenizer block output and queries a full coordinate grid at the target resolution.
For evaluation, we generate 50{,}000 samples and compute FID against the real training set.

\paragraph{Notes on generation on variable geometries.}
Rendering a generated sample requires not only the grouped token representation but also the routing metadata: group centroids $\mu_g \in \mathbb{R}^d$ and extents $\lambda_g \in \mathbb{R}^d_{>0}$.
For uniformly sampled data, the routing metadata is constant across all samples, therefore the diffusion model only needs to generate the grouped token representation, and rendering uses the fixed-grid routing metadata. For data on variable geometries (e.g. on point clouds), the group partition depends on the distribution of observed points for each input, which differs per sample.
The routing metadata must therefore be generated alongside the token representation, e.g.\ by concatenating it to the diffusion target.

\paragraph{Hyperparameters.}
Table~\ref{tab:lhdit_hparams} lists the diffusion model architecture and training configuration for each dataset.

\begin{table}[h]
\centering
\caption{Diffusion model hyperparameters per dataset.
Input structure ($G$, $K$, $C$) is inherited from the tokenizer output representation.}
\label{tab:lhdit_hparams}
\small
\begin{tabular}{lcc}
\toprule
& CIFAR-10 & CelebA-HQ $64^2$ \\
\midrule
\multicolumn{3}{l}{\textit{Input structure (from LH-NeF tok.)}} \\
\quad Groups $G$ & 32 & 64 \\
\quad Tokens/group $K$ & 4 & 2 \\
\quad Token dim $C$ & 16 & 24 \\
\quad Sequence length $L = GK$ & 128 & 128 \\
\midrule
\multicolumn{3}{l}{\textit{Architecture}} \\
\quad Hidden dim (per level) & [384,\,384,\,480,\,576] & [384,\,384,\,480,\,576] \\
\quad Depth (enc/proc/dec blocks) & 3 (7 total) & 3 (7 total) \\
\quad Self-attend layers/block & [1,\,2,\,4,\,4,\,4,\,2,\,1] & [1,\,2,\,4,\,4,\,4,\,2,\,1] \\
\quad Attention heads & 6 & 6 \\
\quad MLP widening factor & 4 & 4 \\
\quad Dropout & 0 & 0 \\
\midrule
\multicolumn{3}{l}{\textit{Diffusion (EDM)}} \\
\quad Parameterization & precond.\ $\hat{\mathbf{z}}_0$ & precond.\ $\hat{\mathbf{z}}_0$ \\
\quad $\sigma_\text{data}$ & $\approx 1$ (measured) & $\approx 1$ (measured) \\
\quad $P_\text{mean}$ / $P_\text{std}$ & $-1.2$ / $1.2$ & $-1.2$ / $1.2$ \\
\quad $\sigma_\text{min}$ / $\sigma_\text{max}$ & $0.002$ / $80$ & $0.002$ / $80$ \\
\quad $\rho$ & 7 & 7 \\
\quad EMA rate & 0.99995 & 0.99995 \\
\midrule
\multicolumn{3}{l}{\textit{Training}} \\
\quad Optimizer & AdamW & AdamW \\
\quad Learning rate & $1.0 \times 10^{-4}$ & $1.53 \times 10^{-4}$ \\
\quad Weight decay & 0.01 & 0 \\
\quad Batch size & 128 & 128 \\
\quad Epochs & 5000 & 6000 \\
\midrule
\multicolumn{3}{l}{\textit{Sampling}} \\
\quad Sampler & Heun (EDM ODE) & Heun (EDM ODE) \\
\quad Steps & 18 & 18 \\
\quad Stochastic churn $S_\text{churn}$ & 0 & 0 \\
\quad Samples for FID & 50{,}000 & 50{,}000 \\
\bottomrule
\end{tabular}
\end{table}

\subsection{Classification}
\label{app:classification}

\paragraph{ShapeNet16.}
The ShapeNet16 classifier is a transformer encoder operating on the tokenization extracted with a frozen LH-NeF tokenizer forward pass and per-channel normalized as in \S\ref{app:generation}. We add two kinds of blocks: \emph{intra-group} blocks self-attend within each group's $K$ tokens, and \emph{inter-group} blocks pool each group to a single vector via mean, self-attend across the $G$ group representations, and broadcast back. This factorization is added for efficiency, with the added benefit of matching the grouped structure of our tokenization. Both block types alternate for a total of 8 layers. We then prepend a learnable CLS token to the per-group pooled representation and apply a 2-layer transformer encoder; the CLS output is fed to a linear classification head. Positional information is injected via random Fourier feature embeddings of the group centers $\mu_g$ and a learned per-slot token-ID embedding to break within-group permutation symmetry.

\begin{table}[h]
\centering
\caption{ShapeNet16 classification hyperparameters.}
\label{tab:classif_shapenet_hparams}
\small
\begin{tabular}{lc}
\toprule
& ShapeNet16 \\
\midrule
$d_\text{model}$ & 256 \\
Depth (blocks) & 8 \\
Attention heads & 8 \\
MLP ratio & 4.0 \\
Dropout & 0.1 \\
Label smoothing & 0.0 \\
Pooling & CLS token \\
Optimizer & AdamW \\
Learning rate & 1.5e-3 \\
Weight decay & 0.29 \\
Batch size & 256 \\
Epochs & 100 \\
\bottomrule
\end{tabular}
\end{table}

\paragraph{CIFAR-10.}
\label{sec:classif_details}

Following the protocol of Spatial Functa~\citep{bauer2023spatial} and ENF~\citep{wessels2024equivariant}, we extract frozen tokenizations on a $50{\times}$-augmented training set (random $32{\times}32$ crops of $40{\times}40$ zero-padded images, horizontal flips) and an unaugmented val/test set, apply per-channel normalization as in \S\ref{app:generation}, and train a classifier on the resulting tokenizations.

The tokenization $\mathbf{Y}^{(L)} \in \mathbb{R}^{G \times K \times C}$ is reshaped into a 2D feature map for convolutions: we order the $G$ groups in row-major order by their centroids $\mu_g$, arrange the $K$ slots within each group as an $H_k \times W_k$ sub-grid, and concatenate to obtain $\mathbf{X} \in \mathbb{R}^{C \times H \times W}$ with $H = H_g H_k$, $W = W_g W_k$ (so $G = H_g W_g$ and $K = H_k W_k$). We then apply a 2-stage ConvNeXt~\citep{liu2022convnet} classifier: a $1{\times}1$ stem to projection dim $D_0$ (no patchify, since $\mathbf{X}$ is already low-resolution), ConvNeXt blocks with stochastic depth, a strided downsampling between stages, global-average pooling, and a linear head. We additionally apply two latent-space augmentations during classifier training: Mixup~\citep{zhang2018mixup} on the feature map and random zeroing of spatial cells. Both are disabled at evaluation.

\begin{table}[h]
\centering
\caption{CIFAR-10 classification hyperparameters (frozen-latent ConvNeXt classifier).}
\label{tab:classif_cifar_hparams}
\small
\begin{tabular}{lc}
\toprule
& CIFAR-10 \\
\midrule
$D_0, D_1$ (per-stage widths) & 192, 384 \\
Blocks per stage & 4, 4 \\
Depthwise kernel size & $3{\times}3$ \\
MLP ratio & 4 \\
Stochastic depth (drop\_path) & 0.2 \\
LayerScale init & $10^{-6}$ \\
Head dropout & 0.1 \\
Group grid $(H_g, W_g)$ & $(4, 8)$ \\
Slot grid $(H_k, W_k)$ & $(2, 2)$ \\
\midrule
Mixup $\alpha$ & 0.5 \\
Token drop $p$ & 0.1 \\
Label smoothing & 0.1 \\
\midrule
Optimizer & AdamW \\
Learning rate (peak) & $1.24 \times 10^{-3}$ \\
Weight decay & $4.15 \times 10^{-2}$ \\
LR schedule & cosine, 1000-iter warmup \\
Batch size & 1024 \\
Epochs & 200 \\
\bottomrule
\end{tabular}
\end{table}

\subsection{ERA5 forecasting}
\label{app:forecasting}

We follow the ENF protocol~\citep{wessels2024equivariant}, where consecutive hourly ERA5 temperature snapshots are tokenized and per-channel normalized (as in \S\ref{app:generation}), and a temporal predictor $F_\psi$ maps $\mathbf{Y}_t \mapsto \mathbf{Y}_{t+1}$ in tokenization space.
Training uses a function-space loss: the predicted tokenization $\hat{\mathbf{Y}}_{t+1} = F_\psi(\mathbf{Y}_t)$ and the ground-truth tokenization $\mathbf{Y}_{t+1}$ are both decoded through the frozen LH-NeF renderer, and the loss is the MSE between the decoded temperature fields.

Our temporal predictor uses the DiT architecture used for generation (\S\ref{app:generation}), with the only difference being the conditioning mechanism: the diffusion model uses adaLN-Zero from a noise-level embedding, while the forecaster uses standard pre-norm (no timestep conditioning, as this is a deterministic prediction task).
The model predicts a \emph{residual} $\Delta\mathbf{Y} = F_\psi(\mathbf{Y}_t)$, and the forecast is $\hat{\mathbf{Y}}_{t+1} = \mathbf{Y}_t + \Delta\mathbf{Y}$.
ENF uses a P$\Theta$NITA MPNN (3 layers, hidden dim 256) as their temporal predictor.

\begin{table}[h]
\centering
\caption{ERA5 forecasting model hyperparameters.}
\label{tab:forecast_hparams}
\small
\begin{tabular}{lcc}
\toprule
& LH-NeF & ENF \\
\midrule
Hidden dim & 256 & 256 \\
Depth (blocks) & 3 & 3 (layers) \\
Self-attention layers per block & 4 & -- \\
Attention heads & 4 & -- \\
Predict residual & yes & yes \\
Optimizer & AdamW & Adam \\
Learning rate & 8.4e-5 & -- \\
Weight decay & 3.9e-3 & -- \\
Batch size & 128 & 32 \\
Epochs & 500 & 1,000 \\
\bottomrule
\end{tabular}
\end{table}

%%%%%%%%%%%%%%%%%%%%%%%%%%%%%%%%%%%%%%%%%%%%%%%%%%%%%%%%%%%%%%%%%%%%%%%%%%%%%%%
%%  PART IV — ADDITIONAL DETAILS AND VISUALIZATIONS
%%%%%%%%%%%%%%%%%%%%%%%%%%%%%%%%%%%%%%%%%%%%%%%%%%%%%%%%%%%%%%%%%%%%%%%%%%%%%%%

\section{Baseline Reproduction}
\label{app:baselines}

Parameter counts (marked with $^\ddagger$ in Table~\ref{tab:reconstruction}) and configurations for the efficiency benchmark are reproduced from reported hyperparameters using the authors' public code where possible.

\paragraph{Functa~\citep{dupont2022data}.}
We use the authors' official JAX/Haiku implementation directly, with the per-dataset hyperparameters reported in their paper.
All Functa models use a SIREN with hidden width 512 and $\omega_0{=}30$.

\begin{table}[h]
\centering
\caption{ENF reproduction hyperparameters per dataset. CIFAR-10 hyperparameters and parameter count (522K) are reported by the authors. For all other datasets, parameter counts are computed from the authors' official JAX implementation using reported hyperparameters.}
\label{tab:enf_hparams}
\small
\begin{tabular}{lcccc}
\toprule
& CelebA-HQ $64^2$ & ImageNet1k $256^2$ & ShapeNet16 & ERA5 \\
\midrule
Hidden dim & 256 & 128 & 128 & 128 \\
Latent dim & 64 & 64 & 32 & 64 \\
Num.\ latents & 36 & 169 & 27 & 36 \\
Num.\ heads & 3$^*$ & 3 & 3 & 3 \\
$k$ (nearest) & 4 & 4 & 4 & 4 \\
$\sigma_q, \sigma_v$ & 2.0, 10.0 & 2.0, 10.0 & 2.0, 10.0 & 2.0, 8.0 \\
\midrule
Params & 3.2M & 817K & 813K & 817K \\
\bottomrule
\end{tabular}
\\[2pt]
{\scriptsize $^*$Not reported for CelebA-HQ; we use the same number of heads as in all other datasets (3).}
\end{table}

\paragraph{Spatial Functa~\citep{bauer2023spatial}.}
Spatial Functa does not provide publicly available code. We use an unofficial public JAX reimplementation~\citep{papa2024spatialfuncta_code} to compute parameter counts. Because this reimplementation is not validated against the authors' original and to avoid reporting inaccurate results, we exclude Spatial Functa from the efficiency benchmark and report only parameter counts.

\begin{table}[h]
\centering
\caption{Functa reproduction hyperparameters per dataset.}
\label{tab:functa_hparams}
\small
\begin{tabular}{lcccc}
\toprule
& CIFAR-10 & CelebA-HQ $64^2$ & ShapeNet16 & ERA5 \\
\midrule
Hidden width & 512 & 512 & 512 & 512 \\
Hidden layers & 10 & 13 & 15 & 16 \\
$\omega_0$ & 30 & 30 & 30 & 30 \\
\midrule
Params & 2.6M & 3.4M & 4.0M & 4.1M \\
\bottomrule
\end{tabular}
\end{table}

\paragraph{ENF~\citep{wessels2024equivariant}.}
For CIFAR-10, the authors report 522K parameters.
For all other datasets, we instantiate ENF's official JAX implementation with the hyperparameters reported in their paper and count parameters directly.

\begin{table}[h]
\centering
\caption{Spatial Functa hyperparameters per dataset.}
\label{tab:sfuncta_hparams}
\small
\begin{tabular}{lcc}
\toprule
& CIFAR-10 & ImageNet1k $256^2$ \\
\midrule
Latent grid & $8{\times}8$ & $32{\times}32$ \\
Latent channels & 16 & 64 \\
Conditioning dims & 1,024 & 65,536 \\
SIREN width & 256 & 256 \\
SIREN layers & 6 & 8 \\
$\omega_0$ & 10 & 10 \\
Modulation & shift-only & shift-only \\
Lat.-to-mod.\ map & 1$\times$1 Conv & 3$\times$3 Conv \\
Interpolation & 1-NN & 1-NN \\
MetaSGD lrs & Yes & Yes \\
\midrule
Params & 425K & 1.4M \\
\bottomrule
\end{tabular}
\end{table}

\section{Efficiency Benchmark Protocol}
\label{app:efficiency_protocol}

Table~\ref{tab:efficiency} benchmarks the cost of a single training step on CelebA-HQ $64^2$.
Measurements are performed on a single NVIDIA H100 NVL GPU (47\,GB MIG partition).
We report parameter count, peak memory at batch size 1 and maximum batch size.

\paragraph{Functa.}
Benchmarked using the authors' official JAX/Haiku implementation~\citep{dupont2022data} with the reported CelebA-HQ $64^2$ hyperparameters (Table~\ref{tab:functa_hparams}).
The training step is a 3-step MAML inner loop ($\eta{=}10^{-2}$ inner-SGD on the per-sample shift modulations) followed by a final SIREN forward pass and an outer-loop backward pass through the SIREN weights, fully fused as a single JIT-compiled XLA program (matching the ENF setup above). Maximum batch size is found via the same binary-search protocol as ENF.

\paragraph{ENF.}
Benchmarked using the authors' official JAX implementation, with JAX 0.9.2 and XLA JIT compilation.
The training step is a 3-step MAML inner loop followed by an outer-loop backward pass through all model parameters (second-order MAML), compiled as a single fused XLA program.
This setup is favorable to ENF: XLA fuses the full inner-loop and outer-gradient computation, eliminating the overhead of retained computation graphs that arises in eager-mode frameworks such as PyTorch.
Maximum batch size is found via binary search over $[1, 4096]$ with 3 validation iterations per candidate.

\paragraph{LH-NeF.}
Benchmarked in PyTorch 2.10 (CUDA 12.8).
The training step is a single forward pass through the tokenizer and renderer followed by a backward pass through all parameters.
Maximum batch size is found via binary search over $[1, 2048]$ with 5 validation iterations per candidate.

\section{Qualitative Results}
\label{app:qualitative}

\subsection{CelebA-HQ $64^2$ generation samples}
\label{app:celeba_samples}

\begin{figure}[h]
\centering
\newlength{\celcolw}\setlength{\celcolw}{0.22\linewidth}%
\begin{tabular*}{\linewidth}{@{\extracolsep{\fill}}cccc@{}}
\includegraphics[width=\celcolw]{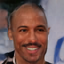} &
\includegraphics[width=\celcolw]{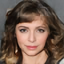} &
\includegraphics[width=\celcolw]{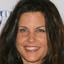} &
\includegraphics[width=\celcolw]{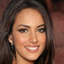} \\[2pt]
\includegraphics[width=\celcolw]{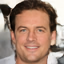} &
\includegraphics[width=\celcolw]{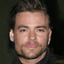} &
\includegraphics[width=\celcolw]{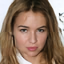} &
\includegraphics[width=\celcolw]{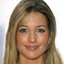} \\[2pt]
\includegraphics[width=\celcolw]{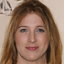} &
\includegraphics[width=\celcolw]{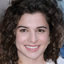} &
\includegraphics[width=\celcolw]{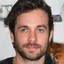} &
\includegraphics[width=\celcolw]{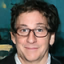} \\
\end{tabular*}
\caption{Additional unconditional generation samples on CelebA-HQ $64^2$ from a diffusion transformer trained on the frozen LH-NeF tokenizations.}
\label{fig:celeba_samples_appx}
\end{figure}

\subsection{ShapeNet16 reconstructions}
\label{app:shapenet_recon}

\begin{figure}[h]
\centering
\newlength{\sncolw}\setlength{\sncolw}{0.126\linewidth}%
\begin{tabular*}{\linewidth}{@{\extracolsep{\fill}}ccccc@{}}
\includegraphics[width=1.2\sncolw]{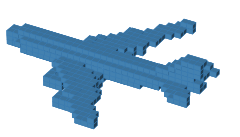} &
\includegraphics[width=\sncolw]{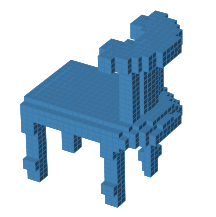} &
\includegraphics[width=\sncolw]{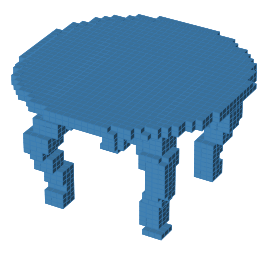} &
\includegraphics[width=\sncolw]{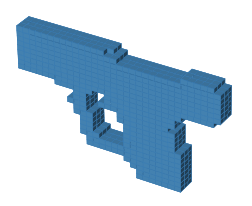} &
\includegraphics[width=\sncolw]{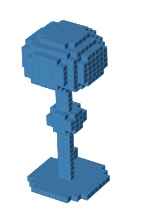} \\[2pt]
\includegraphics[width=1.2\sncolw]{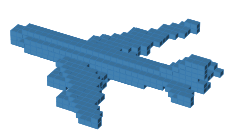} &
\includegraphics[width=0.9\sncolw]{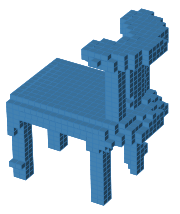} &
\includegraphics[width=\sncolw]{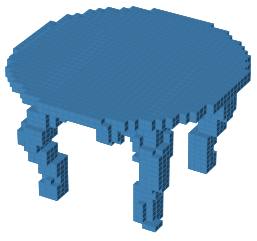} &
\includegraphics[width=\sncolw]{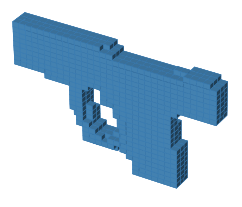} &
\includegraphics[width=0.9\sncolw]{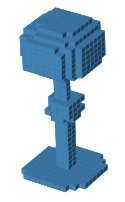} \\[1pt]
{\small Airplane} & {\small Chair} & {\small Table} & {\small Pistol} & {\small Lamp} \\
\end{tabular*}
\caption{ShapeNet16 voxel occupancy reconstructions at $32^3$ resolution.
Top: ground-truth occupancy grids.
Bottom: LH-NeF reconstructions.}
\label{fig:shapenet_recon}
\end{figure}

\clearpage
\section*{Broader Impact}
\label{app:broader_impact}

As a representation learning method that can be paired with generative models on images, it shares the standard misuse risks of image generation systems, e.g. the synthesis of unauthorized or misleading visual content. Positive impacts of modality-agnostic representation learning could include its use in scientific applications. Besides this, we do not anticipate broader societal impacts beyond those typical for foundational representation learning research.

\newpage
\section*{NeurIPS Paper Checklist}

\begin{enumerate}

\item {\bf Claims}
    \item[] Question: Do the main claims made in the abstract and introduction accurately reflect the paper's contributions and scope?
    \item[] Answer: \answerYes{} % Replace by \answerYes{}, \answerNo{}, or \answerNA{}.
    \item[] Justification: Every contribution claimed in the introduction and abstract is well justified by the experiments and derivations. 
    \item[] Guidelines:
    \begin{itemize}
        \item The answer \answerNA{} means that the abstract and introduction do not include the claims made in the paper.
        \item The abstract and/or introduction should clearly state the claims made, including the contributions made in the paper and important assumptions and limitations. A \answerNo{} or \answerNA{} answer to this question will not be perceived well by the reviewers. 
        \item The claims made should match theoretical and experimental results, and reflect how much the results can be expected to generalize to other settings. 
        \item It is fine to include aspirational goals as motivation as long as it is clear that these goals are not attained by the paper. 
    \end{itemize}

\item {\bf Limitations}
    \item[] Question: Does the paper discuss the limitations of the work performed by the authors?
    \item[] Answer: \answerYes{} % Replace by \answerYes{}, \answerNo{}, or \answerNA{}.
    \item[] Justification: Last section in the main body (Section~\ref{sec:limitations}) discusses main limitations and future work directions.
    \item[] Guidelines:
    \begin{itemize}
        \item The answer \answerNA{} means that the paper has no limitation while the answer \answerNo{} means that the paper has limitations, but those are not discussed in the paper. 
        \item The authors are encouraged to create a separate ``Limitations'' section in their paper.
        \item The paper should point out any strong assumptions and how robust the results are to violations of these assumptions (e.g., independence assumptions, noiseless settings, model well-specification, asymptotic approximations only holding locally). The authors should reflect on how these assumptions might be violated in practice and what the implications would be.
        \item The authors should reflect on the scope of the claims made, e.g., if the approach was only tested on a few datasets or with a few runs. In general, empirical results often depend on implicit assumptions, which should be articulated.
        \item The authors should reflect on the factors that influence the performance of the approach. For example, a facial recognition algorithm may perform poorly when image resolution is low or images are taken in low lighting. Or a speech-to-text system might not be used reliably to provide closed captions for online lectures because it fails to handle technical jargon.
        \item The authors should discuss the computational efficiency of the proposed algorithms and how they scale with dataset size.
        \item If applicable, the authors should discuss possible limitations of their approach to address problems of privacy and fairness.
        \item While the authors might fear that complete honesty about limitations might be used by reviewers as grounds for rejection, a worse outcome might be that reviewers discover limitations that aren't acknowledged in the paper. The authors should use their best judgment and recognize that individual actions in favor of transparency play an important role in developing norms that preserve the integrity of the community. Reviewers will be specifically instructed to not penalize honesty concerning limitations.
    \end{itemize}

\item {\bf Theory assumptions and proofs}
    \item[] Question: For each theoretical result, does the paper provide the full set of assumptions and a complete (and correct) proof?
    \item[] Answer: \answerYes{} % Replace by \answerYes{}, \answerNo{}, or \answerNA{}.
    \item[] Justification: The only claims that need proof -- about the invariance of the FiLM modulation -- are derived formally in \cref{app:film_invariance}. All assumptions are clearly stated and formalized in \cref{sec:method}. 
    \item[] Guidelines:
    \begin{itemize}
        \item The answer \answerNA{} means that the paper does not include theoretical results. 
        \item All the theorems, formulas, and proofs in the paper should be numbered and cross-referenced.
        \item All assumptions should be clearly stated or referenced in the statement of any theorems.
        \item The proofs can either appear in the main paper or the supplemental material, but if they appear in the supplemental material, the authors are encouraged to provide a short proof sketch to provide intuition. 
        \item Inversely, any informal proof provided in the core of the paper should be complemented by formal proofs provided in appendix or supplemental material.
        \item Theorems and Lemmas that the proof relies upon should be properly referenced. 
    \end{itemize}

    \item {\bf Experimental result reproducibility}
    \item[] Question: Does the paper fully disclose all the information needed to reproduce the main experimental results of the paper to the extent that it affects the main claims and/or conclusions of the paper (regardless of whether the code and data are provided or not)?
    \item[] Answer: \answerYes{} % Replace by \answerYes{}, \answerNo{}, or \answerNA{}.
    \item[] Justification: Exhaustive experimental information to reproduce our results is provided throughout the Appendix. In addition, results are reproducible with the code provided at the https url in the Introduction. 
    \item[] Guidelines:
    \begin{itemize}
        \item The answer \answerNA{} means that the paper does not include experiments.
        \item If the paper includes experiments, a \answerNo{} answer to this question will not be perceived well by the reviewers: Making the paper reproducible is important, regardless of whether the code and data are provided or not.
        \item If the contribution is a dataset and\slash or model, the authors should describe the steps taken to make their results reproducible or verifiable. 
        \item Depending on the contribution, reproducibility can be accomplished in various ways. For example, if the contribution is a novel architecture, describing the architecture fully might suffice, or if the contribution is a specific model and empirical evaluation, it may be necessary to either make it possible for others to replicate the model with the same dataset, or provide access to the model. In general. releasing code and data is often one good way to accomplish this, but reproducibility can also be provided via detailed instructions for how to replicate the results, access to a hosted model (e.g., in the case of a large language model), releasing of a model checkpoint, or other means that are appropriate to the research performed.
        \item While NeurIPS does not require releasing code, the conference does require all submissions to provide some reasonable avenue for reproducibility, which may depend on the nature of the contribution. For example
        \begin{enumerate}
            \item If the contribution is primarily a new algorithm, the paper should make it clear how to reproduce that algorithm.
            \item If the contribution is primarily a new model architecture, the paper should describe the architecture clearly and fully.
            \item If the contribution is a new model (e.g., a large language model), then there should either be a way to access this model for reproducing the results or a way to reproduce the model (e.g., with an open-source dataset or instructions for how to construct the dataset).
            \item We recognize that reproducibility may be tricky in some cases, in which case authors are welcome to describe the particular way they provide for reproducibility. In the case of closed-source models, it may be that access to the model is limited in some way (e.g., to registered users), but it should be possible for other researchers to have some path to reproducing or verifying the results.
        \end{enumerate}
    \end{itemize}

\item {\bf Open access to data and code}
    \item[] Question: Does the paper provide open access to the data and code, with sufficient instructions to faithfully reproduce the main experimental results, as described in supplemental material?
    \item[] Answer: \answerYes{} % Replace by \answerYes{}, \answerNo{}, or \answerNA{}.
    \item[] Justification: All our results are reproducible through the code provided at the https url in the Introduction; all datasets used in our experiment are publicly available. 
    \item[] Guidelines:
    \begin{itemize}
        \item The answer \answerNA{} means that paper does not include experiments requiring code.
        \item Please see the NeurIPS code and data submission guidelines (\url{https://neurips.cc/public/guides/CodeSubmissionPolicy}) for more details.
        \item While we encourage the release of code and data, we understand that this might not be possible, so \answerNo{} is an acceptable answer. Papers cannot be rejected simply for not including code, unless this is central to the contribution (e.g., for a new open-source benchmark).
        \item The instructions should contain the exact command and environment needed to run to reproduce the results. See the NeurIPS code and data submission guidelines (\url{https://neurips.cc/public/guides/CodeSubmissionPolicy}) for more details.
        \item The authors should provide instructions on data access and preparation, including how to access the raw data, preprocessed data, intermediate data, and generated data, etc.
        \item The authors should provide scripts to reproduce all experimental results for the new proposed method and baselines. If only a subset of experiments are reproducible, they should state which ones are omitted from the script and why.
        \item At submission time, to preserve anonymity, the authors should release anonymized versions (if applicable).
        \item Providing as much information as possible in supplemental material (appended to the paper) is recommended, but including URLs to data and code is permitted.
    \end{itemize}

\item {\bf Experimental setting/details}
    \item[] Question: Does the paper specify all the training and test details (e.g., data splits, hyperparameters, how they were chosen, type of optimizer) necessary to understand the results?
    \item[] Answer: \answerYes{} % Replace by \answerYes{}, \answerNo{}, or \answerNA{}.
    \item[] Justification: Exhaustive experimental information to reproduce our results is provided throughout the Appendix. In addition, results are reproducible with the code provided at the https url in the Introduction.
    \item[] Guidelines:
    \begin{itemize}
        \item The answer \answerNA{} means that the paper does not include experiments.
        \item The experimental setting should be presented in the core of the paper to a level of detail that is necessary to appreciate the results and make sense of them.
        \item The full details can be provided either with the code, in appendix, or as supplemental material.
    \end{itemize}

\item {\bf Experiment statistical significance}
    \item[] Question: Does the paper report error bars suitably and correctly defined or other appropriate information about the statistical significance of the experiments?
    \item[] Answer: \answerYes{} % Replace by \answerYes{}, \answerNo{}, or \answerNA{}.
    \item[] Justification: We report mean $\pm$ standard deviation over 3 independent runs with different random seeds for the main results in \cref{tab:reconstruction}, with the exception of the ImageNet and ERA5 entries. These remaining entries and the downstream task experiments are reported single-seed; full multi-seed results for all experiments will be included in the camera-ready version.
    \item[] Guidelines:
    \begin{itemize}
        \item The answer \answerNA{} means that the paper does not include experiments.
        \item The authors should answer \answerYes{} if the results are accompanied by error bars, confidence intervals, or statistical significance tests, at least for the experiments that support the main claims of the paper.
        \item The factors of variability that the error bars are capturing should be clearly stated (for example, train/test split, initialization, random drawing of some parameter, or overall run with given experimental conditions).
        \item The method for calculating the error bars should be explained (closed form formula, call to a library function, bootstrap, etc.)
        \item The assumptions made should be given (e.g., Normally distributed errors).
        \item It should be clear whether the error bar is the standard deviation or the standard error of the mean.
        \item It is OK to report 1-sigma error bars, but one should state it. The authors should preferably report a 2-sigma error bar than state that they have a 96\% CI, if the hypothesis of Normality of errors is not verified.
        \item For asymmetric distributions, the authors should be careful not to show in tables or figures symmetric error bars that would yield results that are out of range (e.g., negative error rates).
        \item If error bars are reported in tables or plots, the authors should explain in the text how they were calculated and reference the corresponding figures or tables in the text.
    \end{itemize}

\item {\bf Experiments compute resources}
    \item[] Question: For each experiment, does the paper provide sufficient information on the computer resources (type of compute workers, memory, time of execution) needed to reproduce the experiments?
    \item[] Answer: \answerYes{} % Replace by \answerYes{}, \answerNo{}, or \answerNA{}.
    \item[] Justification: Reconstruction stage compute is reported per dataset in Appendix Table~\ref{tab:training_details} (GPU type and count, batch size, training epochs, and approximate wall-clock training time). The training-efficiency benchmark in main text Table~\ref{tab:efficiency} (peak step memory and largest fitting batch size) is performed on a single NVIDIA H100 NVL GPU (47\,GB MIG partition); the protocol is detailed in Appendix~\ref{app:efficiency_protocol}. 
    \item[] Guidelines:
    \begin{itemize}
        \item The answer \answerNA{} means that the paper does not include experiments.
        \item The paper should indicate the type of compute workers CPU or GPU, internal cluster, or cloud provider, including relevant memory and storage.
        \item The paper should provide the amount of compute required for each of the individual experimental runs as well as estimate the total compute. 
        \item The paper should disclose whether the full research project required more compute than the experiments reported in the paper (e.g., preliminary or failed experiments that didn't make it into the paper). 
    \end{itemize}
    
\item {\bf Code of ethics}
    \item[] Question: Does the research conducted in the paper conform, in every respect, with the NeurIPS Code of Ethics \url{https://neurips.cc/public/EthicsGuidelines}?
    \item[] Answer: \answerYes{} % Replace by \answerYes{}, \answerNo{}, or \answerNA{}.
    \item[] Justification: No conflicts with Code of Ethics. 
    \item[] Guidelines:
    \begin{itemize}
        \item The answer \answerNA{} means that the authors have not reviewed the NeurIPS Code of Ethics.
        \item If the authors answer \answerNo, they should explain the special circumstances that require a deviation from the Code of Ethics.
        \item The authors should make sure to preserve anonymity (e.g., if there is a special consideration due to laws or regulations in their jurisdiction).
    \end{itemize}

\item {\bf Broader impacts}
    \item[] Question: Does the paper discuss both potential positive societal impacts and negative societal impacts of the work performed?
    \item[] Answer: \answerYes{} % Replace by \answerYes{}, \answerNo{}, or \answerNA{}.
    \item[] Justification: A brief broader impact discussion is included in Appendix~\ref{app:broader_impact}. As a representation learning method that can be paired with generative models on images, LH-NeF shares the standard misuse risks of image generation systems (e.g., synthesis of unauthorized or misleading visual content). Positive impacts of modality-agnostic representation learning could include its use in scientific applications. Besides this, we do not anticipate broader societal impacts beyond those typical for foundational representation learning research.
    \item[] Guidelines:
    \begin{itemize}
        \item The answer \answerNA{} means that there is no societal impact of the work performed.
        \item If the authors answer \answerNA{} or \answerNo, they should explain why their work has no societal impact or why the paper does not address societal impact.
        \item Examples of negative societal impacts include potential malicious or unintended uses (e.g., disinformation, generating fake profiles, surveillance), fairness considerations (e.g., deployment of technologies that could make decisions that unfairly impact specific groups), privacy considerations, and security considerations.
        \item The conference expects that many papers will be foundational research and not tied to particular applications, let alone deployments. However, if there is a direct path to any negative applications, the authors should point it out. For example, it is legitimate to point out that an improvement in the quality of generative models could be used to generate Deepfakes for disinformation. On the other hand, it is not needed to point out that a generic algorithm for optimizing neural networks could enable people to train models that generate Deepfakes faster.
        \item The authors should consider possible harms that could arise when the technology is being used as intended and functioning correctly, harms that could arise when the technology is being used as intended but gives incorrect results, and harms following from (intentional or unintentional) misuse of the technology.
        \item If there are negative societal impacts, the authors could also discuss possible mitigation strategies (e.g., gated release of models, providing defenses in addition to attacks, mechanisms for monitoring misuse, mechanisms to monitor how a system learns from feedback over time, improving the efficiency and accessibility of ML).
    \end{itemize}
    
\item {\bf Safeguards}
    \item[] Question: Does the paper describe safeguards that have been put in place for responsible release of data or models that have a high risk for misuse (e.g., pre-trained language models, image generators, or scraped datasets)?
    \item[] Answer: \answerNA{} % Replace by \answerYes{}, \answerNo{}, or \answerNA{}.
    \item[] Justification: Our pre-trained checkpoints do not have high risk of misuse, therefore no safeguards are considered.
    \item[] Guidelines:
    \begin{itemize}
        \item The answer \answerNA{} means that the paper poses no such risks.
        \item Released models that have a high risk for misuse or dual-use should be released with necessary safeguards to allow for controlled use of the model, for example by requiring that users adhere to usage guidelines or restrictions to access the model or implementing safety filters. 
        \item Datasets that have been scraped from the Internet could pose safety risks. The authors should describe how they avoided releasing unsafe images.
        \item We recognize that providing effective safeguards is challenging, and many papers do not require this, but we encourage authors to take this into account and make a best faith effort.
    \end{itemize}

\item {\bf Licenses for existing assets}
    \item[] Question: Are the creators or original owners of assets (e.g., code, data, models), used in the paper, properly credited and are the license and terms of use explicitly mentioned and properly respected?
    \item[] Answer: \answerYes{} % Replace by \answerYes{}, \answerNo{}, or \answerNA{}.
    \item[] Justification: Every asset is properly referenced.
    \item[] Guidelines:
    \begin{itemize}
        \item The answer \answerNA{} means that the paper does not use existing assets.
        \item The authors should cite the original paper that produced the code package or dataset.
        \item The authors should state which version of the asset is used and, if possible, include a URL.
        \item The name of the license (e.g., CC-BY 4.0) should be included for each asset.
        \item For scraped data from a particular source (e.g., website), the copyright and terms of service of that source should be provided.
        \item If assets are released, the license, copyright information, and terms of use in the package should be provided. For popular datasets, \url{paperswithcode.com/datasets} has curated licenses for some datasets. Their licensing guide can help determine the license of a dataset.
        \item For existing datasets that are re-packaged, both the original license and the license of the derived asset (if it has changed) should be provided.
        \item If this information is not available online, the authors are encouraged to reach out to the asset's creators.
    \end{itemize}

\item {\bf New assets}
    \item[] Question: Are new assets introduced in the paper well documented and is the documentation provided alongside the assets?
    \item[] Answer: \answerYes{} % Replace by \answerYes{}, \answerNo{}, or \answerNA{}.
    \item[] Justification: All assets are well documented.
    \item[] Guidelines:
    \begin{itemize}
        \item The answer \answerNA{} means that the paper does not release new assets.
        \item Researchers should communicate the details of the dataset\slash code\slash model as part of their submissions via structured templates. This includes details about training, license, limitations, etc. 
        \item The paper should discuss whether and how consent was obtained from people whose asset is used.
        \item At submission time, remember to anonymize your assets (if applicable). You can either create an anonymized URL or include an anonymized zip file.
    \end{itemize}

\item {\bf Crowdsourcing and research with human subjects}
    \item[] Question: For crowdsourcing experiments and research with human subjects, does the paper include the full text of instructions given to participants and screenshots, if applicable, as well as details about compensation (if any)? 
    \item[] Answer: \answerNA{} % Replace by \answerYes{}, \answerNo{}, or \answerNA{}.
    \item[] Justification: Not involved.
    \item[] Guidelines:
    \begin{itemize}
        \item The answer \answerNA{} means that the paper does not involve crowdsourcing nor research with human subjects.
        \item Including this information in the supplemental material is fine, but if the main contribution of the paper involves human subjects, then as much detail as possible should be included in the main paper. 
        \item According to the NeurIPS Code of Ethics, workers involved in data collection, curation, or other labor should be paid at least the minimum wage in the country of the data collector. 
    \end{itemize}

\item {\bf Institutional review board (IRB) approvals or equivalent for research with human subjects}
    \item[] Question: Does the paper describe potential risks incurred by study participants, whether such risks were disclosed to the subjects, and whether Institutional Review Board (IRB) approvals (or an equivalent approval/review based on the requirements of your country or institution) were obtained?
    \item[] Answer: \answerNA{} % Replace by \answerYes{}, \answerNo{}, or \answerNA{}.
    \item[] Justification: Not involved.
    \item[] Guidelines:
    \begin{itemize}
        \item The answer \answerNA{} means that the paper does not involve crowdsourcing nor research with human subjects.
        \item Depending on the country in which research is conducted, IRB approval (or equivalent) may be required for any human subjects research. If you obtained IRB approval, you should clearly state this in the paper. 
        \item We recognize that the procedures for this may vary significantly between institutions and locations, and we expect authors to adhere to the NeurIPS Code of Ethics and the guidelines for their institution. 
        \item For initial submissions, do not include any information that would break anonymity (if applicable), such as the institution conducting the review.
    \end{itemize}

\item {\bf Declaration of LLM usage}
    \item[] Question: Does the paper describe the usage of LLMs if it is an important, original, or non-standard component of the core methods in this research? Note that if the LLM is used only for writing, editing, or formatting purposes and does \emph{not} impact the core methodology, scientific rigor, or originality of the research, declaration is not required.
    %this research? 
    \item[] Answer: \answerYes{} % Replace by \answerYes{}, \answerNo{}, or \answerNA{}.
    \item[] Justification: Declared in the submission.
    \item[] Guidelines:
    \begin{itemize}
        \item The answer \answerNA{} means that the core method development in this research does not involve LLMs as any important, original, or non-standard components.
        \item Please refer to our LLM policy in the NeurIPS handbook for what should or should not be described.
    \end{itemize}

\end{enumerate}

\end{document}